\documentclass[sigconf]{acmart}
\AtBeginDocument{%
  \providecommand\BibTeX{{%
    \normalfont B\kern-0.5em{\scshape i\kern-0.25em b}\kern-0.8em\TeX}}}

\copyrightyear{2023}
\acmYear{2023}
\setcopyright{acmcopyright}
\acmConference[WSDM '23] {Proceedings of the Sixteenth ACM International Conference on Web Search and Data Mining}{February 27--March 3, 2023}{Singapore, Singapore.}
\acmBooktitle{Proceedings of the Sixteenth ACM International Conference on Web Search and Data Mining (WSDM '23), February 27--March 3, 2023, Singapore, Singapore}
\acmPrice{15.00}
\acmISBN{978-1-4503-9407-9/23/02}
\acmDOI{10.1145/3539597.3570446}

\usepackage{xspace}
\usepackage{multirow}
\usepackage{subfigure}
\usepackage{enumitem}
\newcommand{\ourmethod}{{GOOD-D}\xspace}
\newcommand{\ourmethodsimp}{{GOOD-D$_{simp}$}\xspace}

\begin{document}

\title{GOOD-D: On Unsupervised Graph Out-Of-Distribution Detection}


\author{Yixin Liu*}
\affiliation{%
  \institution{Monash University}
  \country{}
}
\email{yixin.liu@monash.edu}

\author{Kaize Ding*}
\affiliation{%
  \institution{Arizona State University}
  \country{}
}
\email{kaize.ding@asu.edu}

\author{Huan Liu}
\affiliation{%
  \institution{Arizona State University}
  \country{}
}
\email{huanliu@asu.edu}

\author{Shirui Pan$\dagger$}
\affiliation{%
  \institution{Griffith University}
  \country{}
}
\email{s.pan@griffith.edu.au}

\makeatletter
\def\authornotetext#1{
	\g@addto@macro\@authornotes{%
	\stepcounter{footnote}\footnotetext{#1}}%
}
\makeatother

\authornotetext{Yixin Liu and Kaize Ding made equal contribution to this work.}
\authornotetext{Shirui Pan is the corresponding author.}



\renewcommand{\shortauthors}{Liu et al.}
\renewcommand{\authors}{Yixin Liu, Kaize Ding, Huan Liu, Shirui Pan}

\begin{abstract}
Most existing deep learning models are trained based on the closed-world assumption, where the test data is assumed to be drawn i.i.d. from the same distribution as the training data, known as in-distribution (ID). However, when models are deployed in an open-world scenario, test samples can be out-of-distribution (OOD) and therefore should be handled with caution. To detect such OOD samples drawn from unknown distribution, OOD detection has received increasing attention lately. However, current endeavors mostly focus on grid-structured data and its application for graph-structured data remains under-explored. Considering the fact that data labeling on graphs is commonly time-expensive and labor-intensive, in this work we study the problem of unsupervised graph OOD detection, aiming at detecting OOD graphs solely based on unlabeled ID data. To achieve this goal, we develop a new graph contrastive learning framework GOOD-D for detecting OOD graphs without using any ground-truth labels. By performing hierarchical contrastive learning on the augmented graphs generated by our perturbation-free graph data augmentation method, GOOD-D is able to capture the latent ID patterns and accurately detect OOD graphs based on the semantic inconsistency in different granularities (i.e., node-level, graph-level, and group-level). As a pioneering work in unsupervised graph-level OOD detection, we build a comprehensive benchmark to compare our proposed approach with different state-of-the-art methods. The experiment results demonstrate the superiority of our approach over different methods on various datasets.\looseness-2
\end{abstract}

\ccsdesc[500]{Mathematics of computing~Graph algorithms}
\ccsdesc[500]{Computing methodologies~Neural networks}

\keywords{Graph Neural Networks, Out-of-distribution Detection, Contrastive Learning}

\maketitle

\section{Introduction}
Nowadays, graphs are ubiquitous in various real-world scenarios, including but not limited to social network analysis~\cite{sage_hamilton2017inductive}, molecular chemistry inference~\cite{drug_wang2021multi}, recommendation~\cite{rs1_yu2021self}, and robotics~\cite{li2020graph}. Expanding deep learning techniques to graph-structured data, graph neural networks (GNNs) have attracted significant research interests in recent years~\cite{gcn_kipf2017semi,gat_velivckovic2018graph,gin_xu2019how}. Based on the message passing scheme, GNNs encode attributive and structural information by feature transformation and message propagation to learn high-level node/graph embeddings, which can be further used for various downstream tasks~\cite{sage_hamilton2017inductive}. 
Attributed to their powerful representation ability and flexibility, GNNs have shown remarkable performance in many graph analytic tasks, such as graph classification~\cite{gin_xu2019how}, link prediction~\cite{zhang2018link}, and node classification~\cite{gcn_kipf2017semi}.

\begin{figure}[t!]
 \centering
 \subfigure[ID and OOD graph samples]{
   \includegraphics[height=2.88cm]{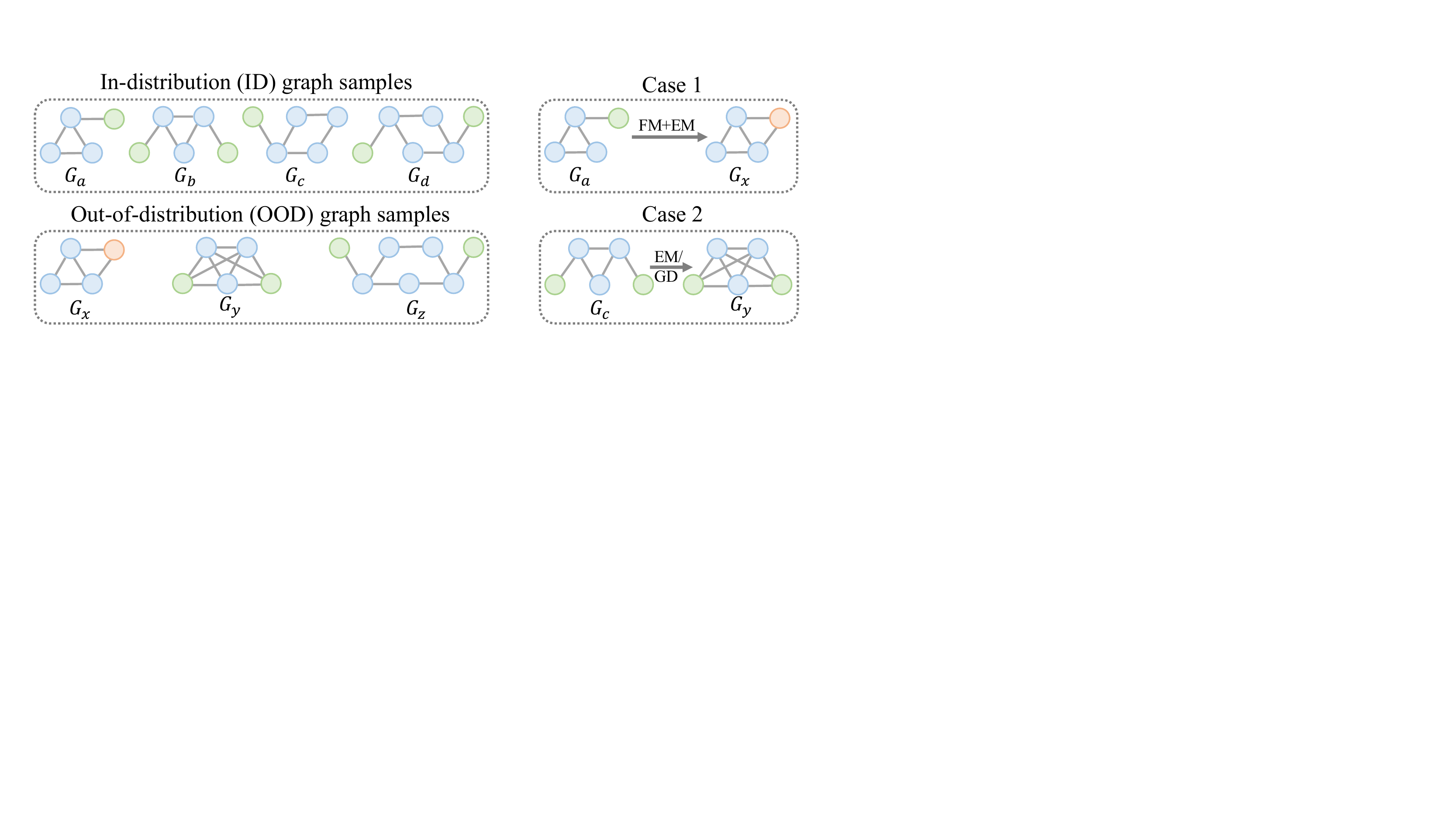}
   \label{subfig:toy1}
 } 
 \hspace{-0.18cm}
 \subfigure[Two example cases]{
   \includegraphics[height=2.88cm]{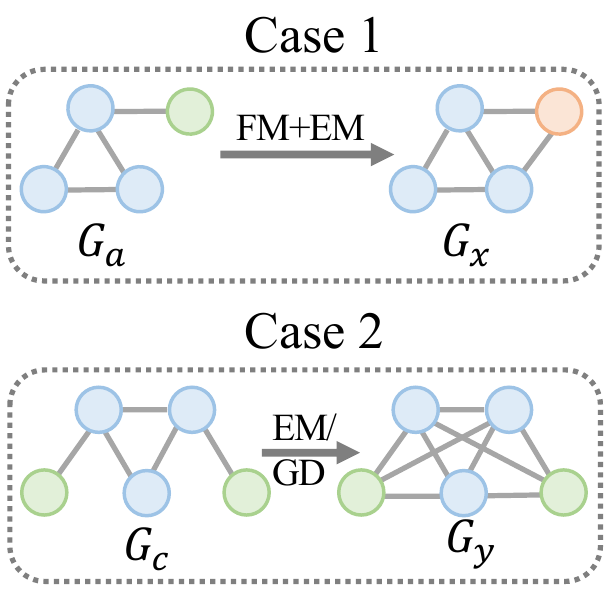}
   \label{subfig:toy2}
 }
 \vspace{-0.5cm}
 \caption{Toy examples of (a) ID graphs ($G_a$ - $G_d$) and OOD graphs ($G_x$ - $G_z$); and (b) perturbation-based augmentations (e.g., feature modification (FM), edge modification (EM), and graph diffusion (GD)) introducing OOD samples.}
 \vspace{-0.5cm}
 \label{fig:toy}
\end{figure}

Despite the prevalence of GNNs for deep graph learning, similar to other modern machine learning models, GNN-based deep graph learning models deployed in the open world often struggle with out-of-distribution (OOD) input samples from a different distribution that the model has not been exposed to during training. Ideally, a reliable machine learning system should not only accurately classify known in-distribution (ID) samples, but also be aware of ``unknown'' OOD inputs during the inference time. This gives rise to the importance of OOD detection, which determines whether an input is ID or OOD and enables the model to take precautions~\cite{supood1_liang2018enhancing,nlpood_zhou2021contrastive,usood_rc_schreyer2017detection}. Recently OOD detection has received increasing attention in images~\cite{ssd_sehwag2020ssd,supood2_hendrycks2016baseline} or text~\cite{nlpood_zhou2021contrastive} domain, while it is substantially less investigated on graph-structured data. Though few recent works~\cite{zhao2020uncertainty,stadler2021graph} in semi-supervised node classification could be used to detect OOD samples, their effectiveness is only confined to node-level detection and largely relies on labeled ID data. It is infeasible to directly apply those methods to detecting OOD graphs, especially when considering the scarcity of class labels and OOD samples. Hence, a natural research question to ask is "\textit{Can we effectively detect OOD graphs solely based on unlabeled in-distribution data?}"

Motivated by the recent progress of self-supervised learning for unsupervised graph representation learning, in this paper, we aim to answer the question by exploring the potential of graph contrastive learning (GCL) for detecting OOD graphs. However, it remains a non-trivial research task, mainly due to the following two reasons: (1) prevailing graph self-supervised learning, especially GCL methods commonly adopt arbitrary augmentations (e.g., feature modification, node/edge dropping, and graph diffusion) to obtain augmented views of the input graph~\cite{ding2022data,gca_zhu2021graph,mvgrl_hassani2020contrastive}. Such augmentations, {as shown in Fig.~\ref{subfig:toy2}}, may unexpectedly perturb both structural and semantic patterns of the graph, which in turn introduces undesired OOD samples~\cite{trans_golan2018deep}. As an example in molecular graphs, perturbing the connection of aspirin might introduce a new molecule with totally different properties, such as five-membered lactone. Hence, proposing a principled \textit{perturbation-free graph augmentation approach} is a necessity of learning expressive graph representations and further detecting OOD samples; (2) Existing GCL methods predominantly focus on instance-level contrast to achieve node/graph-wise discrimination among all the inputs~\cite{gcc_qiu2020gcc,graphcl_you2020graph,gca_zhu2021graph}, which is not well aligned with the objective of OOD detection. 
{As illustrated in Fig.~\ref{subfig:toy1}, in real-world scenarios, OOD graphs may violate the latent patterns of ID graphs in different granularities, such as node-level variation (e.g. $G_x$), graph-level redundant connection (e.g. $G_y$), and cluster-deviated samples (e.g. $G_z$). }
In order to accurately detect diverse OOD graphs during inference, the GCL algorithm is supposed to not only learn expressive node/graph representations based on the augmented graphs, but also consolidate the semantic manifolds (i.e., intra-cluster compactness and inter-cluster separability) of the ID data. Nonetheless, such an unsupervised GCL algorithm as well as the scoring function for detecting OOD graphs have yet to be proposed and investigated.

In this paper, we make the first attempt to solve the problem of unsupervised graph-level OOD detection. 
{To counter the aforementioned challenges, we propose a novel \textbf{\underline{G}}raph  \textbf{\underline{O}}ut-\textbf{\underline{O}}f-\textbf{\underline{D}}istribution \textbf{\underline{D}}etection method, namely \ourmethod. Our theme is to capture the latent patterns shared by ID graphs via hierarchical contrastive learning with perturbation-free data augmentation.} 
To address \textit{the first challenge}, we design a perturbation-free data augmentation method to enable graph self-supervised learning without introducing detrimental perturbations. Specifically, we generate a structure view of the input graph by rewriting the features of each node with the pre-computed high-level structural encodings. By maximizing the agreement between the representations learned from the structure view and the original graph (i.e., feature view), the model will learn to extract consistent representations from the different views of an ID graph. To address \textit{the second challenge}, we propose a hierarchical graph contrastive learning algorithm, which not only enables node and graph-level contrasts to learn expressive node and graph representations, but also incorporates group-level contrast to enhance the semantic manifold of the ID data. Thus for each test graph sample, its node-level and graph-level disagreement between two different views as well as the group-level disconfirmation to the ID data semantic manifold can be leveraged as an indicative OOD scoring function. To automatically control the contribution of the hierarchical contrastive learning at each granularity, we further equip the hierarchical contrastive learning component with an adaptive learning loss. Finally, we construct a comprehensive benchmark for graph-level OOD detection based on real-world datasets from diverse domains to evaluate the effectiveness of our proposed framework against state-of-the-art methods. Based upon it, we conduct extensive experiments to demonstrate the superiority of our approach. In summary, our major contributions are three-fold: \looseness-2

\begin{itemize}[leftmargin=*,noitemsep,topsep=1.5pt]

    \item \textbf{Problem:} We formally formulate the graph-level OOD detection problem and build a set of benchmarking datasets for evaluation, which can shed good light on the following research in this field.
    
    \item \textbf{Algorithm:} We propose a self-supervised graph OOD detection framework, i.e., \ourmethod, which can learn expressive ID distribution and measure the OOD scores for different inputs by performing hierarchical contrastive learning with perturbation-free graph data augmentation.
    
    \item \textbf{Evaluations:} We conduct extensive experiments on a range of benchmarks to demonstrate the superior performance of \ourmethod over the state-of-the-art methods. 
    
\end{itemize}

\section{Related Work}
\noindent\textbf{Graph Neural Networks.} 
Graph neural networks (GNNs) have attracted increasing research attention due to their capability to model graph-structured data~\cite{gcn_kipf2017semi, gat_velivckovic2018graph, sage_hamilton2017inductive, gin_xu2019how, sgc_wu2019simplifying}. A branch of methods termed spectral-based GNNs defines graph convolution based on spectral graph theory~\cite{cheb_defferrard2016convolutional, gcn_kipf2017semi}. For example, GCN~\cite{gcn_kipf2017semi} performs convolutional operation via the first-order approximation of Chebyshev polynomial filter~\cite{cheb_defferrard2016convolutional}. SGC~\cite{sgc_wu2019simplifying} further simplifies the graph convolution to a linear operation. 
Another family of models termed spatial-based GNNs performs graph convolution by aggregating and transforming local information~\cite{sage_hamilton2017inductive,gat_velivckovic2018graph,gin_xu2019how}. For instance, GAT~\cite{gat_velivckovic2018graph} introduces the attention mechanism to allocate weights for neighbors in local aggregation. GIN~\cite{gin_xu2019how} boosts the expressive power of GNNs by utilizing an injective summation operation to aggregate neighboring information. Some recent works try to improve from different perspectives, including scalability~\cite{saint_zeng2019graphsaint}, trustworthy~\cite{zhang2022trustworthy}, and architecture 
design~\cite{mrgnas_zheng2022mrgnas}. \looseness-2

\noindent\textbf{Out-of-distribution Detection. } 
Out-of-distribution (OOD) detection aims to discriminate the test samples that are far from the distribution of training samples. According to the availability of ground-truth labels during training phase, we can divide OOD detection methods into two types, i.e., supervised methods and unsupervised methods~\cite{ssd_sehwag2020ssd,ngc_wu2021ngc}. Supervised methods~\cite{supood1_liang2018enhancing,supood2_hendrycks2016baseline} leverage fine-grained labels to model the distribution of in-distribution (ID) data detect the OOD samples in the learned feature space. Unsupervised methods capture the distribution of ID data via reconstruction-based models~\cite{usood_rc_schreyer2017detection}, one-class classification~\cite{oc_ruff2018deep}, probabilistic models~\cite{pm_ren2019likelihood}, and self-supervised learning~\cite{ssd_sehwag2020ssd,nlpood_zhou2021contrastive}. Considering the expensive cost of label annotation \cite{ssd_sehwag2020ssd}, this paper investigates unsupervised OOD detection, which is a more practical but also more challenging scenario compared to the supervised counterpart.

While extensive OOD detection methods are developed for vision~\cite{ssd_sehwag2020ssd,ngc_wu2021ngc,supood1_liang2018enhancing} and language~\cite{nlpood_zhou2021contrastive} data, how to identify OOD samples on graph-structured data is still under-explored. 
There is a line of studies~\cite{oodg1_li2022out,oodg3_fan2021generalizing} aim to generalize GNNs to OOD data under distribution shifts. However, these methods focus on improving the generalization ability of GNNs on certain downstream tasks (e.g., node classification) rather than identifying the OOD samples. 
{Another related research topic is graph anomaly detection, which can be regarded as a subfield of OOD detection~\cite{dominant_ding2019deep,ding2021inductive,ocgin_zhao2021using,cola_liu2021anomaly}. Graph anomaly detection focuses on detecting malicious data from real-world systems (e.g., fraud or spam data)~\cite{comga_luo2022comga} or the tail samples belonging to minority categories~\cite{glocalkd_ma2022deep}. By contrast, graph OOD detection is a more general and challenging task, since malicious/tail samples can be regarded as the subtypes of OOD data~\cite{ssd_sehwag2020ssd}. }
In this paper, we consider several anomaly detection methods~\cite{ocgin_zhao2021using,glocalkd_ma2022deep} for comparison, and also verify the effectiveness our method on both OOD detection and anomaly detection tasks. 

\noindent\textbf{Graph Contrastive Learning. } 
As an important branch of graph self-supervised learning~\cite{ssl_survey_liu2021self,gssl_survey_liu2022graph}, graph contrastive learning (GCL) has shown to be an effective technique for unsupervised graph representation learning~\cite{dgi_velickovic2019deep,infograph_sun2020infograph,mvgrl_hassani2020contrastive,graphcl_you2020graph,gca_zhu2021graph,gcc_qiu2020gcc,ggd_zheng2022rethinking,ugcl_zheng2022unifying,ding2022structural}. A general pipeline of GCL methods is to generate multiple graph views via data augmentation and then maximize the cross-view mutual agreement between samples with similar semantics~\cite{graphcl_you2020graph,gca_zhu2021graph,mvgrl_hassani2020contrastive,ugcl_zheng2022unifying,ding2022structural}. Apart from representation learning, GCL also benefits various graph-related applications, such as recommendation systems~\cite{rs1_yu2021self}, drug interaction learning~\cite{drug_wang2021multi}, and graph structure learning~\cite{sublime_liu2022towards}. In this paper, we apply GCL to graph-level OOD detection tasks by innovatively equipping GCL with structure-based 
perturbation-free augmentation and hierarchical contrast.

\section{Problem Definition}
Before formulating the research problem, we first provide some necessary notations. Let $G=(\mathcal{V},\mathcal{E},\mathbf{X})$ represent a graph, where $\mathcal{V}$ is the set of nodes and $\mathcal{E}$ is the set of edges. The node features are represented by the feature matrix $\mathbf{X} \in \mathbb{R}^{n \times d_f}$, where $n=|\mathcal{V}|$ is the number of nodes and $d_f$ is the feature dimension. The structure information can also be described by an adjacency matrix $\mathbf{A} \in \mathbb{R}^{n \times n}$, so a graph can be alternatively represented by $G=(\mathbf{A},\mathbf{X})$. 

In this paper, we focus on the unsupervised graph-level out-of-distribution (OOD) detection problem, which can be formulated as: \looseness-2

\begin{definition}[Unsupervised graph-level OOD detection] 
We assume that we have an ID dataset $\mathcal{D}^{in}=\{G_1^{in}, \cdots\, G_{N_1}^{in}\}$ where graphs are sampled from a certain distribution $\mathbb{P}^{in}$ and an OOD dataset $\mathcal{D}^{out}=\{G_1^{out}, \cdots\, G_{N_2}^{out}\}$ where graphs are sampled from an OOD distribution $\mathbb{P}^{out}$. Given a graph $G$, the goal is to correctly identify its source distribution (i.e., $\mathbb{P}^{in}$ or $\mathbb{P}^{out}$). Concretely, a scoring function $f(\cdot)$ is learned to generate an OOD detection score $s=f(G)$ for an input graph $G$, where a larger $s$ indicates a higher probability that $G$ is from $\mathbb{P}^{out}$. In practice, the scoring function (i.e. learning model) is trained only on ID dataset $\mathcal{D}^{in}_{train} \subset \mathcal{D}^{in}$ and is evaluated on a test set containing $\mathcal{D}^{in}_{test} \subset \mathcal{D}^{in}$  ($\mathcal{D}^{in}_{test} \cap \mathcal{D}^{in}_{train} = \emptyset$) and $\mathcal{D}^{out}_{test} \subset \mathcal{D}^{out}$.
\end{definition}

Note that graph data from $\mathbb{P}^{in}$ and $\mathbb{P}^{out}$ may belong to one or more categories. Since we investigate the unsupervised OOD problem, all the category-based labels are not considered.

\section{Methodology}
\begin{figure*}[!ht]
  \includegraphics[width=0.9\textwidth]{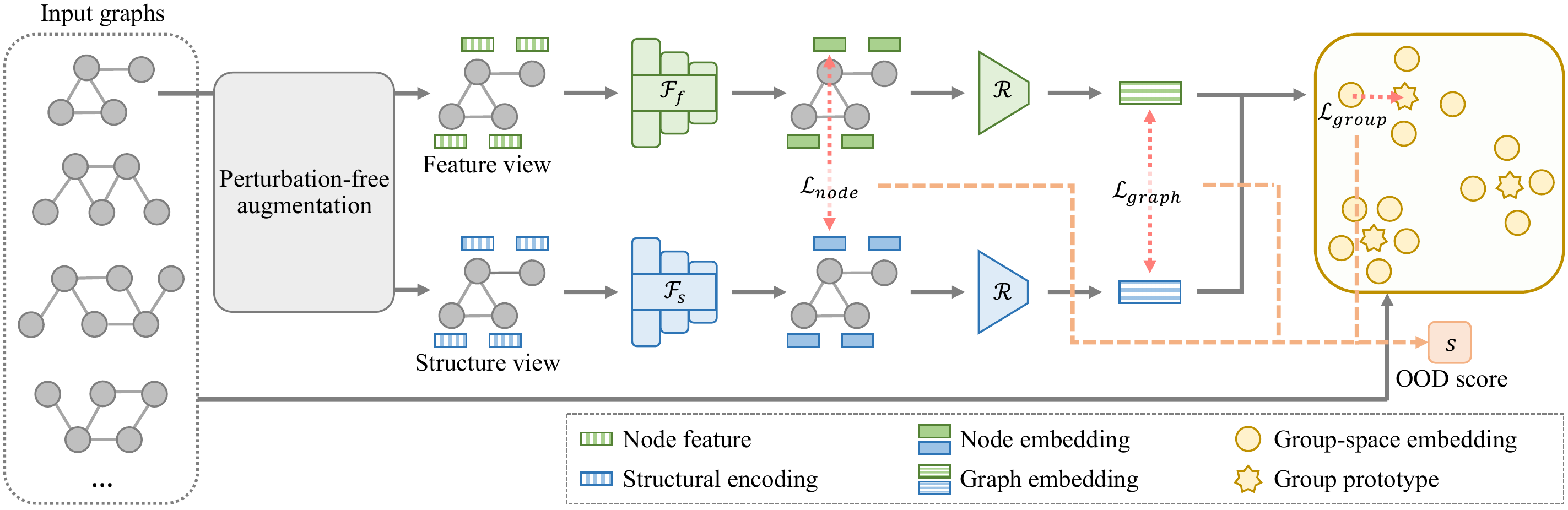}
  \vspace{-4mm}
  \caption{An overall illustration of the proposed method \ourmethod.}
  \label{fig:pipeline}
  \vspace{-3mm}
\end{figure*}

In this section, we introduce our proposed novel \textbf{\underline{G}}raph  \textbf{\underline{O}}ut-\textbf{\underline{O}}f-\textbf{\underline{D}}istribution \textbf{\underline{D}}etection (\ourmethod) method. The overall pipeline of \ourmethod is illustrated in Fig. \ref{fig:pipeline}. {For each input graph sample, we first construct feature view and structure view via \textit{perturbation-free graph data augmentation}. Then, node embeddings and graph embeddings are generated by two GNN-based encoders ($\mathcal{F}_{f}$ and $\mathcal{F}_{s}$) and readout functions ($\mathcal{R}$). After that, we conduct \textit{hierarchical contrastive learning} at three different levels, i.e., node level, graph level, and group level, which maximizes the intra- and inter- graph agreement from multiple perspectives. Finally, the OOD detection score $s$ is estimated by an adaptive scoring mechanism that aggregates the contrastive errors at three levels.} In the following sub-sections, we will introduce the design of \ourmethod in detail.

\vspace{-1.5mm}
\subsection{\hspace{-0.1cm}Perturbation-free Graph Data Augmentation} \label{subsec:view}

The core of contrastive learning is to maximize the agreement between samples in two different views~\cite{simclr_chen2020simple}. To construct views with different contents, a general solution is data augmentation, i.e., generating views with stochastic data transformation~\cite{moco_he2020momentum}. For graph data, conventional augmentations are mainly based on random data perturbation, such as edge perturbation~\cite{dgi_velickovic2019deep}, node dropping~\cite{graphcl_you2020graph}, subgraph extraction~\cite{gcc_qiu2020gcc}, graph diffusion~\cite{mvgrl_hassani2020contrastive}, and feature modification~\cite{gca_zhu2021graph}. Through maximizing the agreement between a graph and its augmented view, the GCL model can learn high-quality representations which are invariant to the perturbations~\cite{graphcl_you2020graph}. \looseness-2

{Although such perturbation invariance is usually conducive to representation learning, it may not always benefit OOD detection. The data perturbation on graphs, unexpectedly, can create undesired OOD graphs from original ID graphs, since the ID and OOD data are sometimes similar and can be transformed to each other with few modifications~\cite{bai2019simgnn}. Guided by the objective of contrastive learning, transformed graphs are enforced to have similar embeddings to the original ones, making the model less sensitive to the difference between ID data and potential OOD data~\cite{trans_golan2018deep}. In this case, the perturbation-based data augmentations would deteriorate rather than boost the performance of OOD detection.}

{To address the above issue, we propose a perturbation-free graph augmentation strategy specialized in contrastive OOD detection. Following our new augmentation strategy, two fixed, distinct, and informative views are constructed from node features and graph structure perspectives respectively.} Specifically, given a graph $G$, the \textbf{feature view} is directly built by integrating the node features and adjacency matrix, i.e., $G_{f} = (\mathbf{A}, \mathbf{X})$. To construct the \textbf{structure view}, we extract \textit{node-level structural encodings} from the graph structure and combine them with adjacency matrix, i.e., $G_{s} = (\mathbf{A}, \mathbf{S})$, where $\mathbf{S} \in \mathbb{R}^{n \times d_s}$ is a structural encoding matrix and each row $\mathbf{s}_i$ indicates a $d_s$-dimensional structural encoding vector that incorporates structure-related properties of the corresponding node $v_i$.

To capture universal topological patterns from graph structure, we jointly consider \textit{global} and \textit{local} structural information when generating structural encodings. To capture global structural information, we use a random walk diffusion process to build global structural encodings~\cite{lspe_dwivedi2022graph}. Concretely, the encoding $\mathbf{s}_i^{(rw)}$ of node $v_i$ can be acquired by collecting the diagonal elements of multi-step random walk-based graph diffusion matrices:

\vspace{-3.0mm}
\begin{equation}
\vspace{-1.8mm}
\label{eq:rw_enc}
\mathbf{s}_i^{(rw)}=\left[{\mathbf{T}}_{i i}, {\mathbf{T}}_{i i}^{2}, \cdots, {\mathbf{T}}_{i i}^{d_s^{(rw)}}\right] \in \mathbb{R}^{d_s^{(rw)}},
\vspace{-0.0mm}
\end{equation}

\noindent where ${\mathbf{T}}=\mathbf{A}\mathbf{D}^{-1}$ is the random walk transition matrix, $\mathbf{D}$ is the diagonal degree matrix such that $\mathbf{D}_{i i}=\sum_{j} \mathbf{A}_{i j}$, and $d_s^{(rw)}$ is the dimension of random walk-based global structural encodings. Attributed to the characteristic of graph diffusion, the global encodings represent the unique global role (e.g., central node or tail node) for each node. 
To capture local structural information, we define local structural encodings as the one-hot vector of node degrees~\cite{gin_xu2019how,gcc_qiu2020gcc}: 

\vspace{-3.0mm}
\begin{equation}
\vspace{-1.8mm}
\label{eq:dg_enc}
\mathbf{s}_{ik}^{(dg)}=\left\{
\begin{aligned}
1,  & \text{ $k = \mathbf{D}_{i i}$ or $k = d_s^{(dg)} < \mathbf{D}_{i i}$} \\
0,  & \text{ $k \neq \mathbf{D}_{i i} $}
\end{aligned}
\xspace,
\right.
\vspace{-0.0mm}
\end{equation}

\noindent where $\mathbf{s}_{ik}^{(dg)}$ is the $k$-th element of degree-based local structure encoding vector $\mathbf{s}_i^{(dg)}$ for node $v_i$ and $d_s^{(dg)}$ is the dimension of degree-based local structural encodings. The degree indicates the popularity of each node, representing its local role from a neighboring subgraph. 
Finally, the structural encoding is acquired by concatenating the global and local encodings, i.e., $\mathbf{s}_i = [\mathbf{s}_i^{(rw)}|| \mathbf{s}_i^{(dg)}]$. It is worth noting that our approach is agnostic to the definition of structural encoding, meaning that diverse structural encodings (such as distance~\cite{dispe_you2019position} and Laplacian eigenvectors~\cite{lappe_dwivedi2020benchmarking}) can be applied to \ourmethod. We leave this technical extension for future works.

\vspace{-1.5mm}
\subsection{Hierarchical Graph Contrastive Learning} \label{subsec:hcl}

Given two graph views $G_{f}$ and $G_{s}$, our proposed hierarchical contrastive learning model first extracts node embeddings and graph embeddings with \textit{GNN encoders and readout function}, and then conducts hierarchical contrastive learning at three different levels, i.e., node level, graph level, and group level. 

\subsubsection{GNN encoders and readout function}

To effectively extract informative node embeddings from two graph views, we utilize two parallel GNN encoders (denoted as \textit{feature-view encoder} $\mathcal{F}_f$ and \textit{structure-view encoder} $\mathcal{F}_s$) for representation learning. Different from most GCL frameworks with weight-shared encoders~\cite{graphcl_you2020graph,gca_zhu2021graph,gcc_qiu2020gcc}, in \ourmethod, the weights of $\mathcal{F}_f$ and $\mathcal{F}_s$ are independent to each other. The reason is that feature view and structure view have different contents and input feature spaces, and it would be harmful to encode distinct information with the same set of parameters. 

Considering its powerful expression ability, we employ GIN~\cite{gin_xu2019how} ($\epsilon=0$ for simplicity) as GNN encoders. Taking $\mathcal{F}_f$ as an example, the propagation rule in the $l$-th layer of GIN can be expressed as:

\vspace{-3.0mm}
\begin{equation}
\vspace{-1.8mm}
\label{eq:gin}
\mathbf{h}_{i}^{(f,l)}=\operatorname{MLP}^{(f,l)}\left(\mathbf{h}_{i}^{(f,l-1)}+\sum_{v_j \in \mathcal{N}(v_i)} \mathbf{h}_{j}^{(f,l-1)}\right),
\vspace{-0.0mm}
\end{equation}

\noindent where $\mathbf{h}_{i}^{(f,l)}$ is the interval embedding of node $v_i$ at the $l$-th layer of feature-view encoder $\mathcal{F}_f$, $\mathcal{N}(v_i)$ is the set of first-order neighborhood nodes of node $v_i$, and $\operatorname{MLP}$ is a two-layer multi-layer perceptron (MLP) network. We set $\mathbf{h}_{i}^{(f,0)} = \mathbf{x}_{i}$ in $\mathcal{F}_f$ and $\mathbf{h}_{i}^{(s,0)} = \mathbf{s}_{i}$ in $\mathcal{F}_s$. Given an $L$-layer $\mathcal{F}_f$, the final feature-view node embedding of node $v_i$ is acquired by concatenating the interval embeddings at each layer, i.e., $\mathbf{h}_{i}^{(f)} = [\mathbf{h}_{i}^{(f,1)}|| \cdots ||\mathbf{h}_{i}^{(f,L)}]$, and we can compute structure-view node embedding $\mathbf{h}_{i}^{(s)}$ in the same way. 

After we get the node embeddings, we use a readout function to acquire the graph embedding. Following GIN~\cite{gin_xu2019how}, we employ summation as our readout function, which can be represented by:

\vspace{-3.0mm}
\begin{equation}
\vspace{-1.8mm}
\label{eq:readout}
\mathbf{h}_{G}^{(f)}=\sum_{v_i \in \mathcal{V}_G} \mathbf{h}_{i}^{(f)},\quad  \mathbf{h}_{G}^{(s)}=\sum_{v_i \in \mathcal{V}_G} \mathbf{h}_{i}^{(s)},
\vspace{-0.0mm}
\end{equation}

\noindent where $\mathbf{h}_{G}^{(f)}$ and $\mathbf{h}_{G}^{(s)}$ is the feature- and structure- view graph embedding of input graph $G$ respectively, and $\mathcal{V}_G$ is the node set of $G$. \looseness-2

\subsubsection{Hierarchical Graph Contrastive Learning}

{Our core idea is to capture the common patterns of training ID data through contrastive learning, such that the OOD data samples that violate these patterns can be easily exposed during inference.  
Most existing GCL methods conduct contrast at a single scale level, e.g., node level~\cite{gca_zhu2021graph}, subgraph level~\cite{gcc_qiu2020gcc}, and graph level~\cite{graphcl_you2020graph}. Some GCL methods leverage cross-level contrast~\cite{dgi_velickovic2019deep,mvgrl_hassani2020contrastive} to extract inter-scale knowledge within a graph. Despite their success in representation learning, these methods may suffer from sub-optimal OOD detection performance due to the following misalignment. 
Firstly, existing GCL methods mainly employ instance-level discrimination, ignoring the intra-cluster compactness and inter-cluster separability of ID data. However, such semantic manifolds are significant for OOD detection, since OOD samples usually appear as cluster-deviated samples of ID data~\cite{oodg1_li2022out}. Moreover, most GCL methods conduct contrastive learning at a single scale, while the distinguishable graph patterns exist at multiple levels due to the diversity of OOD data~\cite{glocalkd_ma2022deep}.} \looseness-2 

{To overcome these shortages, we propose a novel hierarchical contrastive learning method for graph OOD detection. To model semantic manifolds of ID data, we establish a group-level contrast mechanism that maximizes the agreement between each sample and its clustering prototype. To capture the patterns at multiple levels, we conduct contrastive learning at three different levels, i.e., node level, graph level, and group level.}

\textbf{Node-level contrast} aims to find the intrinsic patterns from the perspective of nodes within a single graph. To this end, the learning objective is to maximize the agreement between the embeddings belonging to the same node on two views. To conduct contrast in a specific latent space, we first map $\mathbf{h}_{i}^{(f)}$ and $\mathbf{h}_{i}^{(s)}$ into node-space embeddings $\mathbf{z}_{i}^{(f)}$ and $\mathbf{z}_{i}^{(s)}$ with MLP-based projection networks. After that, an InfoNCE-like~\cite{simclr_chen2020simple,gca_zhu2021graph} node-level contrastive loss is built to maximize the node-level agreement:

\vspace{-3.0mm}
\begin{equation}
\vspace{-1.1mm}
\label{eq:loss_node}
\begin{aligned}
\mathcal{L}_{node} = \frac{1}{|\mathcal{B}|} \sum_{G_j \in \mathcal{B}} \frac{1}{2|\mathcal{V}_{G_j}|} \sum_{v_i \in \mathcal{V}_{G_j}} \Big[ \ell(\mathbf{z}_{i}^{(f)},\mathbf{z}_{i}^{(s)}) + \ell(\mathbf{z}_{i}^{(s)},\mathbf{z}_{i}^{(f)}) \Big], \\
\ell(\mathbf{z}_{i}^{(f)},\mathbf{z}_{i}^{(s)}) = - \operatorname{log}\frac{e^{\operatorname{sim}(\mathbf{z}_{i}^{(f)},\mathbf{z}_{i}^{(s)})/\tau}}{\sum_{v_k \in \mathcal{V}_{G_j} \backslash v_i} e^{\operatorname{sim}(\mathbf{z}_{i}^{(f)},\mathbf{z}_{k}^{(s)})/\tau} },
\end{aligned}
\vspace{-0.0mm}
\end{equation}

\noindent where $\mathcal{B}$ is a training batch containing multiple graph samples, $\mathcal{V}_{G_j}$ is the node set of graph $G_j$, $\operatorname{sim}(\cdot,\cdot)$ is the cosine similarity function, $\tau$ is the temperature parameter, $\ell(\mathbf{z}_{i}^{(s)},\mathbf{z}_{i}^{(f)})$ is calculated following $\ell(\mathbf{z}_{i}^{(f)},\mathbf{z}_{i}^{(s)})$.

\textbf{Graph-level contrast} focuses on modeling the cross-view agreement on each graph sample. Similar to node-level contrast, the graph embeddings $\mathbf{h}_{G}^{(f)}$ and $\mathbf{h}_{G}^{(s)}$ are transformed into graph-space embeddings $\mathbf{z}_{G}^{(f)}$ and $\mathbf{z}_{G}^{(s)}$ with MLP-based projection networks. Then, we construct a graph-level contrastive loss for mutual agreement maximization:

\vspace{-3.0mm}
\begin{equation}
\vspace{-1.8mm}
\label{eq:loss_graph}
\begin{aligned}
\mathcal{L}_{graph} = \frac{1}{2|\mathcal{B}|} \sum_{G_i \in \mathcal{B}} \Big[ \ell(\mathbf{z}_{G_i}^{(f)},\mathbf{z}_{G_i}^{(s)}) + \ell(\mathbf{z}_{G_i}^{(s)},\mathbf{z}_{G_i}^{(f)}) \Big], \\
\ell(\mathbf{z}_{G_i}^{(f)},\mathbf{z}_{G_i}^{(s)}) = - \operatorname{log}\frac{e^{\operatorname{sim}(\mathbf{z}_{G_i}^{(f)},\mathbf{z}_{G_i}^{(s)})/\tau}}{\sum_{G_j \in \mathcal{B} \backslash G_i} e^{\operatorname{sim}(\mathbf{z}_{G_i}^{(f)},\mathbf{z}_{G_j}^{(s)})/\tau} },
\end{aligned}
\vspace{-0.0mm}
\end{equation}

\noindent where $\ell(\mathbf{z}_{G_i}^{(s)},\mathbf{z}_{G_i}^{(f)})$ is calculated following $\ell(\mathbf{z}_{G_i}^{(f)},\mathbf{z}_{G_i}^{(s)})$, and other notations are similar to Eq. (\ref{eq:loss_node}). 

\textbf{Group-level contrast} targets to capture the patterns shared by a group of graph samples. To this end, we first perform clustering algorithm to find prototypes~\cite{fedproto_tan2022fedproto,pcl_li2020prototypical,fedpcl_tan2022federated} (cluster centroids) and use prototypical contrastive learning loss to maximize the agreement between each sample and its corresponding prototype. Specifically, for each graph $G_i$, we first concatenate $\mathbf{h}_{G_i}^{(f)}$ with $\mathbf{h}_{G_i}^{(s)}$, and project it into a group-space embedding $\mathbf{z}_{G_i}$. At the beginning of each epoch, we perform k-means clustering over all group-space embeddings and allocate prototype for each sample. Based on the prototypes $\mathcal{C} = \{\mathbf{c}_i\}_{i=1}^{K}$ defined as the average group-space embedding of each cluster, the group-level contrastive loss can be calculated by: 

\vspace{-3.0mm}
\begin{equation}
\vspace{-1.8mm}
\label{eq:loss_group}
\mathcal{L}_{group} = - \frac{1}{|\mathcal{B}|} \sum_{G_i \in \mathcal{B}} \operatorname{log}\frac{e^{\operatorname{sim}(\mathbf{z}_{G_i},\mathbf{c}_{j})/\tau_j}}{\sum_{c_k \in \mathcal{C} \backslash c_j} e^{\operatorname{sim}(\mathbf{z}_{G_i},\mathbf{c}_{k})/\tau_k} }, 
\vspace{-0.0mm}
\end{equation}

\noindent where $c_j$ is the prototype corresponding to graph sample $G_i$, $\tau_j$ and $\tau_k$ are the concentration level-based temperatures~\cite{pcl_li2020prototypical} that have positive correlation with the squared deviation of the $j$-th and $k$-th clusters, respectively. {Intuitively, the cluster number $K$ should be highly related to the class distribution of ID data. Although this label information is unknown in unsupervised settings, we empirically find that \ourmethod works well with a moderate $K$ value owing to its low sensitivity to this hyper-parameter (see Sec.~\ref{subsec:param}). }

To learn the shared patterns at different levels simultaneously, the hierarchical contrastive learning model is optimized by jointly minimizing the above three loss functions:

\vspace{-3.0mm}
\begin{equation}
\vspace{-1.8mm}
\label{eq:loss}
\mathcal{L} = \mathcal{L}_{node} + \mathcal{L}_{graph} + \mathcal{L}_{group}.
\vspace{-0.0mm}
\end{equation}

\subsection{Adaptive Training and OOD Scoring} \label{subsec:scoring}

\noindent \textbf{Error-based OOD scoring.}
Through optimizing the loss function (Eq. (\ref{eq:loss})), \ourmethod is able to capture the regularity information of ID graph data at node, graph, and group levels. That is to say, given an ID graph sample as input, the predicted error is expected to be small, indicating the latent patterns of this sample highly match the learned ones. Motivated by this, we calculate the OOD score based on the predicted errors of testing samples. To be concrete, for an input graph $G_i$, the node-level OOD score $s_{G_i}^{(node)}$ and graph-level OOD score $s_{G_i}^{(graph)}$ are obtained by computing the node-level and graph-level contrastive losses of this sample, respectively. In group-level contrast, we do not perform the clustering algorithm on testing data but allocate groups by selecting the closest prototype to its group-space embedding. Then, the group-level OOD score $s_{G_i}^{(group)}$ is computed based on the similarity of prototype and group-space embedding. In the simple version of \ourmethod, the OOD score is the summation of the scores of three levels:

\vspace{-3.0mm}
\begin{equation}
\vspace{-1.8mm}
\label{eq:score}
s_{G_i} = s_{G_i}^{(node)} + s_{G_i}^{(graph)} + s_{G_i}^{(group)}.
\vspace{-0.0mm}
\end{equation}

\begin{table*}[t]
\centering
\caption{OOD detection results in terms of AUC (in percent, mean $\pm$ std). The best and runner-up results are highlighted with \textbf{bold} and \underline{underline}, respectively.} 
\label{tab:main_result_od}
\vspace{-0.3cm}
\resizebox{1\textwidth}{!}{
\begin{tabular}{l | cccccccccc|c}
\toprule
ID dataset & BZR & PTC-MR & AIDS & ENZYMES & IMDB-M & Tox21 & FreeSolv & BBBP & ClinTox & Esol & \multirow{2}{1.8em}{Avg. Rank}  \\
\cline{1-11}
OOD dataset & COX2 & MUTAG & DHFR & PROTEIN & IMDB-B & SIDER & ToxCast & BACE & LIPO & MUV \\
\hline
PK-LOF      & $42.22{\scriptstyle\pm8.39}$ & $51.04{\scriptstyle\pm6.04}$ & $50.15{\scriptstyle\pm3.29}$ & $50.47{\scriptstyle\pm2.87}$ & $48.03{\scriptstyle\pm2.53}$ & $51.33{\scriptstyle\pm1.81}$ & $49.16{\scriptstyle\pm3.70}$ & $53.10{\scriptstyle\pm2.07}$ & $50.00{\scriptstyle\pm2.17}$ & $50.82{\scriptstyle\pm1.48}$ & $11.9$ \\
PK-OCSVM    & $42.55{\scriptstyle\pm8.26}$ & $49.71{\scriptstyle\pm6.58}$ & $50.17{\scriptstyle\pm3.30}$ & $50.46{\scriptstyle\pm2.78}$ & $48.07{\scriptstyle\pm2.41}$ & $51.33{\scriptstyle\pm1.81}$ & $48.82{\scriptstyle\pm3.29}$ & $53.05{\scriptstyle\pm2.10}$ & $50.06{\scriptstyle\pm2.19}$ & $51.00{\scriptstyle\pm1.33}$ & $11.8$ \\
PK-iF & $51.46{\scriptstyle\pm1.62}$ & $54.29{\scriptstyle\pm4.33}$ & $51.10{\scriptstyle\pm1.43}$ & $51.67{\scriptstyle\pm2.69}$ & $50.67{\scriptstyle\pm2.47}$ & $49.87{\scriptstyle\pm0.82}$ & $52.28{\scriptstyle\pm1.87}$ & $51.47{\scriptstyle\pm1.33}$ & $50.81{\scriptstyle\pm1.10}$ & $50.85{\scriptstyle\pm3.51}$ & $10.1$ \\
WL-LOF   &   $48.99{\scriptstyle\pm6.20}$ & $53.31{\scriptstyle\pm8.98}$ & $50.77{\scriptstyle\pm2.87}$ & $52.66{\scriptstyle\pm2.47}$ & $52.28{\scriptstyle\pm4.50}$ & $51.92{\scriptstyle\pm1.58}$ & $51.47{\scriptstyle\pm4.23}$ & $52.80{\scriptstyle\pm1.91}$ & $51.29{\scriptstyle\pm3.40}$ & $51.26{\scriptstyle\pm1.31}$ & $9.3$ \\
WL-OCSVM    & $49.16{\scriptstyle\pm4.51}$ & $53.31{\scriptstyle\pm7.57}$ & $50.98{\scriptstyle\pm2.71}$ & $51.77{\scriptstyle\pm2.21}$ & $51.38{\scriptstyle\pm2.39}$ & $51.08{\scriptstyle\pm1.46}$ & $50.38{\scriptstyle\pm3.81}$ & $52.85{\scriptstyle\pm2.00}$ & $50.77{\scriptstyle\pm3.69}$ & $50.97{\scriptstyle\pm1.65}$ & $10.0$ \\
WL-iF  & $50.24{\scriptstyle\pm2.49}$ & $51.43{\scriptstyle\pm2.02}$ & $50.10{\scriptstyle\pm0.44}$ & $51.17{\scriptstyle\pm2.01}$ & $51.07{\scriptstyle\pm2.25}$ & $50.25{\scriptstyle\pm0.96}$ & $52.60{\scriptstyle\pm2.38}$ & $50.78{\scriptstyle\pm0.75}$ & $50.41{\scriptstyle\pm2.17}$ & $50.61{\scriptstyle\pm1.96}$ & $11.3$ \\
\hline
InfoGraph-iF & $63.17{\scriptstyle\pm9.74}$ & $51.43{\scriptstyle\pm5.19}$ & $93.10{\scriptstyle\pm1.35}$ & $60.00{\scriptstyle\pm1.83}$ & $58.73{\scriptstyle\pm1.96}$ & $56.28{\scriptstyle\pm0.81}$ & $56.92{\scriptstyle\pm1.69}$ & $53.68{\scriptstyle\pm2.90}$ & $48.51{\scriptstyle\pm1.87}$ & $54.16{\scriptstyle\pm5.14}$ & $7.4$ \\
InfoGraph-MD & $86.14{\scriptstyle\pm6.77}$ & $50.79{\scriptstyle\pm8.49}$ & $69.02{\scriptstyle\pm11.67}$ & $55.25{\scriptstyle\pm3.51}$ & $\mathbf{81.38{\scriptstyle\pm1.14}}$ & $59.97{\scriptstyle\pm2.06}$ & $58.05{\scriptstyle\pm5.46}$ & $70.49{\scriptstyle\pm4.63}$ & $48.12{\scriptstyle\pm5.72}$ & $77.57{\scriptstyle\pm1.69}$ & $6.5$ \\
GraphCL-iF & $60.00{\scriptstyle\pm3.81}$ & $50.86{\scriptstyle\pm4.30}$ & $92.90{\scriptstyle\pm1.21}$ & $61.33{\scriptstyle\pm2.27}$ & $59.67{\scriptstyle\pm1.65}$ & $56.81{\scriptstyle\pm0.97}$ & $55.55{\scriptstyle\pm2.71}$ & $59.41{\scriptstyle\pm3.58}$ & $47.84{\scriptstyle\pm0.92}$ & $62.12{\scriptstyle\pm4.01}$ & $7.7$ \\
GraphCL-MD & $83.64{\scriptstyle\pm6.00}$ & $73.03{\scriptstyle\pm2.38}$ & $93.75{\scriptstyle\pm2.13}$ & $52.87{\scriptstyle\pm6.11}$ & $79.09{\scriptstyle\pm2.73}$ & $58.30{\scriptstyle\pm1.52}$ & $60.31{\scriptstyle\pm5.24}$ & $75.72{\scriptstyle\pm1.54}$ & $51.58{\scriptstyle\pm3.64}$ & $78.73{\scriptstyle\pm1.40}$ & $4.3$ \\
\hline
OCGIN  & $76.66{\scriptstyle\pm4.17}$ & $\underline{80.38{\scriptstyle\pm6.84}}$ & $86.01{\scriptstyle\pm6.59}$ & $57.65{\scriptstyle\pm2.96}$ & $67.93{\scriptstyle\pm3.86}$ & $46.09{\scriptstyle\pm1.66}$ & $59.60{\scriptstyle\pm4.78}$ & $61.21{\scriptstyle\pm8.12}$ & $49.13{\scriptstyle\pm4.13}$ & $54.04{\scriptstyle\pm5.50}$ & $6.9$ \\
GLocalKD  & $75.75{\scriptstyle\pm5.99}$ & $70.63{\scriptstyle\pm3.54}$ & $93.67{\scriptstyle\pm1.24}$ & $57.18{\scriptstyle\pm2.03}$ & $78.25{\scriptstyle\pm4.35}$ & $\underline{66.28{\scriptstyle\pm0.98}}$ & $64.82{\scriptstyle\pm3.31}$ & $73.15{\scriptstyle\pm1.26}$ & $55.71{\scriptstyle\pm3.81}$ & $86.83{\scriptstyle\pm2.35}$ & $4.1$ \\
\hline 
\ourmethodsimp & $\underline{93.00{\scriptstyle\pm3.20}}$ & $78.43{\scriptstyle\pm2.67}$ & $\underline{98.91{\scriptstyle\pm0.41}}$ & $\mathbf{61.89{\scriptstyle\pm2.51}}$ & $79.71{\scriptstyle\pm1.19}$ & $65.30{\scriptstyle\pm1.27}$ & $\underline{70.48{\scriptstyle\pm2.75}}$ & $\underline{81.56{\scriptstyle\pm1.97}}$ & $\underline{66.13{\scriptstyle\pm2.98}}$ & $\underline{91.39{\scriptstyle\pm0.46}}$ & $\underline{2.2}$ \\
\ourmethod & $\mathbf{94.99{\scriptstyle\pm2.25}}$ & $\mathbf{81.21{\scriptstyle\pm2.65}}$ & $\mathbf{99.07{\scriptstyle\pm0.40}}$ & $\underline{61.84{\scriptstyle\pm1.94}}$ & $\underline{79.94{\scriptstyle\pm1.09}}$ & $\mathbf{66.50{\scriptstyle\pm1.35}}$ & $\mathbf{80.13{\scriptstyle\pm3.43}}$ & $\mathbf{82.91{\scriptstyle\pm2.58}}$ & $\mathbf{69.18{\scriptstyle\pm3.61}}$ & $\mathbf{91.52{\scriptstyle\pm0.70}}$ & $\mathbf{1.2}$ \\
\bottomrule
\end{tabular}}
\vspace{-0.2cm}
\end{table*}

\noindent \textbf{Adaptive training and scoring.}
By adding the loss terms (via Eq. (\ref{eq:loss}))  at three levels, we can easily train an OOD detection model; we can also obtain the OOD scores of the testing data based on the predicted errors at different levels, as defined Eq. (\ref{eq:score}). 
However, treating three terms equally would ignore the diverse sensitivities at different levels, leading to sub-optimal performance. On the one hand, different ID datasets may have their distinctive shared patterns at different graph scale levels; on the other hand, it is not trivial to manually tune the trade-off weights among three training and testing terms, especially in unsupervised scenarios. To alleviate this issue, we design an adaptive training and scoring mechanism that automatically allocates the weights for loss and score terms. 

Concretely, in training phase, we introduce the standard deviations of predicted errors to balance the loss terms of different levels. The adaptive loss function is computed by:

\vspace{-3.0mm}
\begin{equation}
\vspace{-1.8mm}
\label{eq:loss_adp}
\mathcal{L} = (\sigma_{node})^{\alpha} \mathcal{L}_{node} + (\sigma_{graph})^{\alpha} \mathcal{L}_{graph} + (\sigma_{group})^{\alpha} \mathcal{L}_{group},
\vspace{-0.0mm}
\end{equation}

\noindent where $\sigma_{node}$, $\sigma_{graph}$ and $\sigma_{group}$ are the standard deviations of predicted errors of the corresponding levels, and $\alpha \geq 0$ is a hyper-parameter that controls the strength of self-adaptiveness. Our motivation is to punish the loss term with a larger deviation, thus our model can better concentrate on capturing the shared patterns at the corresponding level. 

In inference phase, to balance the scores of different levels, we employ z-score normalization based on the mean values and standard deviations of the predicted errors of training samples:

\vspace{-3.0mm}
\begin{equation}
\label{eq:score_adp}
\vspace{-0.6mm}
s_{G_i} = \frac{s_{G_i}^{(node)} - \mu_{node}}{\sigma_{node}} + \frac{s_{G_i}^{(graph)} - \mu_{graph}}{\sigma_{graph}} + \frac{s_{G_i}^{(group)} - \mu_{group}}{\sigma_{group}},
\vspace{-0.0mm}
\end{equation}

\noindent where $\mu_{node}$, $\mu_{graph}$, and $\mu_{group}$ are the mean values of predicted errors of training samples of the corresponding levels. The intuition is to normalize the score of each level with ID data and thus highlight the OOD samples with a larger bias. Notably, $\mu$ and $\sigma$ can be computed based on the losses in the last epoch, requiring no additional computational complexity. 

\vspace{-1.5mm}
\subsection{Complexity Analysis}

We analyze the time complexity of each component in \ourmethod. For data augmentation, the structural encoding can be calculated at once during pre-processing phase, and the computational complexities of random walk encoding and degree encoding are $\mathcal{O}(Nnmd^{(rw)}_s)$ and $\mathcal{O}(Nn)$ respectively, where $N$ is the number of graphs, $n$ is the (average) number of nodes, and $m$ is the (average) number of edges. For GNN encoders, the complexity is $\mathcal{O}(NLmd_h + NLnd_h^2 + Nnd_h(d_f + d_s))$, where $d_h$, $d_f$, and $d_s$ are the dimension of interval embedding, raw feature, and structural encoding, respectively. For three contrastive learning modules, the complexity of each $L'$-layer projection network is $\mathcal{O}(NL'nd_p^2)$, where $d_p=Ld_h$ is the dimension of projected embedding. The complexities of contrastive losses are $\mathcal{O}(Nn^2d_p)$, $\mathcal{O}(Nbd_p)$ and $\mathcal{O}(NKd_p)$ respectively, where $b$ is the batch size. The $I$-iter k-means clustering further brings $\mathcal{O}(IKNd_p)$ complexity. In the inference phase, the OOD scoring has a computational cost similar to the forward propagation in each training step. The adaptive mechanism does not cost extra computation. Thus, after ignoring the smaller terms, the overall complexity of each training epoch is $\mathcal{O}(NLd_h(m + n^2 + b + IK + nLL'd_h) + Nnd_h(d_f+d_s))$.

\section{Experiments}
In this section, we empirically evaluate the effectiveness of the proposed \ourmethod framework. In particular, the experiments are unfolded by answering the following research questions:

\begin{itemize}[leftmargin=*,noitemsep,topsep=1.5pt]
    \item \textbf{RQ1:} How effective is \ourmethod on identifying OOD graphs?
    \item \textbf{RQ2:} Can we apply \ourmethod to anomaly detection tasks?
    \item \textbf{RQ3:} What are the contributions of contrastive learning at different levels in \ourmethod framework?
    \item \textbf{RQ4:} Is \ourmethod sensitive to the hyper-parameters?
    \item \textbf{RQ5:} What kind of embeddings in each space and OOD score distribution are learned by \ourmethod?
\end{itemize}

\begin{table*}[h]
\centering
\caption{Anomaly detection results in terms of AUC (in percent, mean $\pm$ std). The best and runner-up results are highlighted with \textbf{bold} and \underline{underline}, respectively.} 
\label{tab:main_result_ad}
\vspace{-0.3cm}
\resizebox{1\textwidth}{!}{
\begin{tabular}{l | cccc|cc|cc|cc}
\toprule
Method & PK-OCSVM & PK-iF & WL-OCSVM & WL-iF & InfoGraph-iF & GraphCL-iF & OCGIN & GLocalKD & \ourmethodsimp & \ourmethod \\
\hline
PROTEINS-full & $50.49{\scriptstyle\pm4.92}$ & $60.70{\scriptstyle\pm2.55}$ & $51.35{\scriptstyle\pm4.35}$ & $61.36{\scriptstyle\pm2.54}$ & $57.47{\scriptstyle\pm3.03}$ & $60.18{\scriptstyle\pm2.53}$ & $70.89{\scriptstyle\pm2.44}$ & $\mathbf{77.30{\scriptstyle\pm5.15}}$ & $\underline{74.74{\scriptstyle\pm2.28}}$ & $71.97{\scriptstyle\pm3.86}$ \\
ENZYMES       & $53.67{\scriptstyle\pm2.66}$ & $51.30{\scriptstyle\pm2.01}$ & $55.24{\scriptstyle\pm2.66}$ & $51.60{\scriptstyle\pm3.81}$ & $53.80{\scriptstyle\pm4.50}$ & $53.60{\scriptstyle\pm4.88}$ & $58.75{\scriptstyle\pm5.98}$ & $\underline{61.39{\scriptstyle\pm8.81}}$ & $61.23{\scriptstyle\pm4.58}$ & $\mathbf{63.90{\scriptstyle\pm3.69}}$ \\
AIDS          & $50.79{\scriptstyle\pm4.30}$ & $51.84{\scriptstyle\pm2.87}$ & $50.12{\scriptstyle\pm3.43}$ & $61.13{\scriptstyle\pm0.71}$ & $70.19{\scriptstyle\pm5.03}$ & $79.72{\scriptstyle\pm3.98}$ & $78.16{\scriptstyle\pm3.05}$ & $93.27{\scriptstyle\pm4.19}$ & $\underline{94.09{\scriptstyle\pm1.75}}$ & $\mathbf{97.28{\scriptstyle\pm0.69}}$ \\
DHFR          & $47.91{\scriptstyle\pm3.76}$ & $52.11{\scriptstyle\pm3.96}$ & $50.24{\scriptstyle\pm3.13}$ & $50.29{\scriptstyle\pm2.77}$ & $52.68{\scriptstyle\pm3.21}$ & $51.10{\scriptstyle\pm2.35}$ & $49.23{\scriptstyle\pm3.05}$ & $56.71{\scriptstyle\pm3.57}$ & $\mathbf{62.71{\scriptstyle\pm3.38}}$ & $\underline{62.67{\scriptstyle\pm3.11}}$ \\
BZR           & $46.85{\scriptstyle\pm5.31}$ & $55.32{\scriptstyle\pm6.18}$ & $50.56{\scriptstyle\pm5.87}$ & $52.46{\scriptstyle\pm3.30}$ & $63.31{\scriptstyle\pm8.52}$ & $60.24{\scriptstyle\pm5.37}$ & $65.91{\scriptstyle\pm1.47}$ & $69.42{\scriptstyle\pm7.78}$ & $\underline{74.48{\scriptstyle\pm4.91}}$ & $\mathbf{75.16{\scriptstyle\pm5.15}}$ \\
COX2          & $50.27{\scriptstyle\pm7.91}$ & $50.05{\scriptstyle\pm2.06}$ & $49.86{\scriptstyle\pm7.43}$ & $50.27{\scriptstyle\pm0.34}$ & $53.36{\scriptstyle\pm8.86}$ & $52.01{\scriptstyle\pm3.17}$ & $53.58{\scriptstyle\pm5.05}$ & $59.37{\scriptstyle\pm12.67}$ & $\underline{60.46{\scriptstyle\pm12.34}}$ & $\mathbf{62.65{\scriptstyle\pm8.14}}$ \\
DD            & $48.30{\scriptstyle\pm3.98}$ & $71.32{\scriptstyle\pm2.41}$ & $47.99{\scriptstyle\pm4.09}$ & $70.31{\scriptstyle\pm1.09}$ & $55.80{\scriptstyle\pm1.77}$ & $59.32{\scriptstyle\pm3.92}$ & $72.27{\scriptstyle\pm1.83}$ & $\mathbf{80.12{\scriptstyle\pm5.24}}$ & $72.24{\scriptstyle\pm1.82}$ & $\underline{73.25{\scriptstyle\pm3.19}}$ \\
NCI1          & $49.90{\scriptstyle\pm1.18}$ & $50.58{\scriptstyle\pm1.38}$ & $50.63{\scriptstyle\pm1.22}$ & $50.74{\scriptstyle\pm1.70}$ & $50.10{\scriptstyle\pm0.87}$ & $49.88{\scriptstyle\pm0.53}$ & $\mathbf{71.98{\scriptstyle\pm1.21}}$ & $\underline{68.48{\scriptstyle\pm2.39}}$ & $59.56{\scriptstyle\pm1.62}$ & $61.12{\scriptstyle\pm2.21}$ \\
IMDB-B        & $50.75{\scriptstyle\pm3.10}$ & $50.80{\scriptstyle\pm3.17}$ & $54.08{\scriptstyle\pm5.19}$ & $50.20{\scriptstyle\pm0.40}$ & $56.50{\scriptstyle\pm3.58}$ & $56.50{\scriptstyle\pm4.90}$ & $60.19{\scriptstyle\pm8.90}$ & $52.09{\scriptstyle\pm3.41}$ & $\underline{65.49{\scriptstyle\pm1.06}}$ & $\mathbf{65.88{\scriptstyle\pm0.75}}$ \\
REDDIT-B      & $45.68{\scriptstyle\pm2.24}$ & $46.72{\scriptstyle\pm3.42}$ & $49.31{\scriptstyle\pm2.33}$ & $48.26{\scriptstyle\pm0.32}$ & $68.50{\scriptstyle\pm5.56}$ & $71.80{\scriptstyle\pm4.38}$ & $75.93{\scriptstyle\pm8.65}$ & $77.85{\scriptstyle\pm2.62}$ & $\underline{87.87{\scriptstyle\pm1.38}}$ & $\mathbf{88.67{\scriptstyle\pm1.24}}$ \\
COLLAB        & $49.59{\scriptstyle\pm2.24}$ & $50.49{\scriptstyle\pm1.72}$ & $52.60{\scriptstyle\pm2.56}$ & $50.69{\scriptstyle\pm0.32}$ & $46.27{\scriptstyle\pm0.73}$ & $47.61{\scriptstyle\pm1.29}$ & $60.70{\scriptstyle\pm2.97}$ & $52.94{\scriptstyle\pm0.85}$ & $\underline{62.10{\scriptstyle\pm0.63}}$ & $\mathbf{72.08{\scriptstyle\pm0.90}}$ \\
HSE           & $57.02{\scriptstyle\pm8.42}$ & $56.87{\scriptstyle\pm10.51}$ & $62.72{\scriptstyle\pm10.13}$ & $53.02{\scriptstyle\pm5.12}$ & $53.56{\scriptstyle\pm3.98}$ & $51.18{\scriptstyle\pm2.71}$ & $64.84{\scriptstyle\pm4.70}$ & $59.48{\scriptstyle\pm1.44}$ & $\underline{69.18{\scriptstyle\pm1.89}}$ & $\mathbf{69.65{\scriptstyle\pm2.14}}$ \\
MMP           & $46.65{\scriptstyle\pm6.31}$ & $50.06{\scriptstyle\pm3.73}$ & $55.24{\scriptstyle\pm3.26}$ & $52.68{\scriptstyle\pm3.34}$ & $54.59{\scriptstyle\pm2.01}$ & $54.54{\scriptstyle\pm1.86}$ & $\mathbf{71.23{\scriptstyle\pm0.16}}$ & $67.84{\scriptstyle\pm0.59}$ & $70.18{\scriptstyle\pm1.14}$ & $\underline{70.51{\scriptstyle\pm1.56}}$ \\
p53           & $46.74{\scriptstyle\pm4.88}$ & $50.69{\scriptstyle\pm2.02}$ & $54.59{\scriptstyle\pm4.46}$ & $50.85{\scriptstyle\pm2.16}$ & $52.66{\scriptstyle\pm1.95}$ & $53.29{\scriptstyle\pm2.32}$ & $58.50{\scriptstyle\pm0.37}$ & $\underline{64.20{\scriptstyle\pm0.81}}$ & $\mathbf{66.48{\scriptstyle\pm0.56}}$ & $62.99{\scriptstyle\pm1.55}$ \\
PPAR-gamma    & $53.94{\scriptstyle\pm6.94}$ & $45.51{\scriptstyle\pm2.58}$ & $57.91{\scriptstyle\pm6.13}$ & $49.60{\scriptstyle\pm0.22}$ & $51.40{\scriptstyle\pm2.53}$ & $50.30{\scriptstyle\pm1.56}$ & $\mathbf{71.19{\scriptstyle\pm4.28}}$ & $64.59{\scriptstyle\pm0.67}$ & $66.85{\scriptstyle\pm2.19}$ & $\underline{67.34{\scriptstyle\pm1.71}}$ \\
\hline
Avg. Rank    & $8.7$ & $7.7$ & $6.9$ & $7.5$ & $6.5$ & $6.9$ & $3.5$ & $3.1$ & $\underline{2.3}$ & $\mathbf{1.7}$  \\
\bottomrule
\end{tabular}}
\vspace{-0.2cm}
\end{table*}

\vspace{-1.5mm}
\subsection{Experimental Settings}

\subsubsection{Datasets}
Previous studies on OOD detection mainly focus on image or language datasets, while few investigate OOD detection on graph datasets. In this paper, inspired by existing studies~\cite{nlpood_zhou2021contrastive,usood_rc_schreyer2017detection}, we establish a benchmark for graph-level OOD detection by using different pairs of graph datasets as ID and OOD data, respectively. 
We select 10 pairs of datasets from two mainstream graph data benchmarks (i.e., TU datasets~\cite{tu_Morris2020} and OGB~\cite{ogb_hu2020open}), where datasets in each pair belong to the same field and have moderate domain shift. We select 8 pairs of molecule datasets, 1 pair of bioinformatics datasets, and 1 pair of social network datasets. $90\%$ of ID samples are used for training, and $10\%$ of ID samples and the same number of OOD samples are integrated together for testing. 
We also conduct experiments on anomaly detection settings, where 15 datasets from TU benchmark~\cite{tu_Morris2020} are used for evaluation. Following the setting in~\cite{glocalkd_ma2022deep}, the samples in minority class or real anomalous class are viewed as anomalies, while the rest are viewed as normal data. Similar to~\cite{glocalkd_ma2022deep,ocgin_zhao2021using}, only normal data are used for model training.

\subsubsection{Baselines}
We compare \ourmethod and \ourmethodsimp (i.e., \ourmethod without adaptive training and scoring) with baseline approaches in the following three categories:

\noindent \textbf{Graph kernel+detector.} This type of methods first extracts vectorized representations by graph kernels~\cite{gk_vishwanathan2010graph}, and uses OOD/anomaly detectors to identify OOD samples based on representations. We take Weisfeiler-Lehman kernel (WL)~\cite{wlgk_shervashidze2011weisfeiler} and propagation kernel (PK)~\cite{pk_neumann2016propagation} as kernels, and take local outlier factor (LOF)~\cite{lof_breunig2000lof}, one-class SVM (OCSVM)~\cite{ocsvm_manevitz2001one}, and isolation forest (iF)~\cite{iforest_liu2008isolation} as detectors. 

\noindent \textbf{GCL+detector.} This type of methods generates representations with state-of-the-art GCL methods, and discriminates OOD samples with detectors based on learned representations. We select two graph-level GCL methods (i.e., InfoGraph~\cite{infograph_sun2020infograph} and GraphCL~\cite{graphcl_you2020graph}) for representation learning. Apart from iF detector~\cite{iforest_liu2008isolation}, we also consider Mahalanobis distance-based (MD) detector which is proved to be effective for detecting OOD data~\cite{ssd_sehwag2020ssd,nlpood_zhou2021contrastive}.

\noindent \textbf{End-to-end.} We compare our method with two graph anomaly detection methods which are trained in an end-to-end manner. The first method is OCGIN~\cite{ocgin_zhao2021using}, where a GIN encoder is optimized with a SVDD objective. The second method is GLocalKD~\cite{glocalkd_ma2022deep} which identifies anomalies via knowledge distillation.

\subsubsection{Evaluation and Implementation}

We evaluate our method using a popular OOD detection metric, i.e., area under receiver operating characteristic Curve (AUC). Higher AUC values indicate better detection performance. We conduct all experiments by repeating 5 times and report the mean AUC and standard deviation. 
We perform grid search to select the key hyper-parameters of \ourmethod. For all baselines, we also use the optimal parameter settings from the corresponding papers or obtained by grid search. 
The \textbf{code and more implementation details} are available at \url{https://github.com/yixinliu233/G-OOD-D}.

\vspace{-1.5mm}
\subsection{Performance on OOD Detection (RQ1)} \label{subsec:od_results}

To answer RQ1, we compare our proposed methods with 12 competing methods. The AUC results are reported in Table~\ref{tab:main_result_od}. From the comparison results, we make the following observations. 1) \ourmethod outperforms all baselines on 8 groups of datasets and achieves runner-up performance on the rest of datasets. Meanwhile, our proposed method has the best average rank across all compared methods. These results demonstrate the effectiveness of \ourmethod in detecting OOD samples from various graph-structured data. 2) Compared to methods except for \ourmethod, \ourmethodsimp also achieves very competitive results, indicating an average rank of 2.2. The results illustrate that equally considering the contrastive learning in three levels is also powerful in OOD detection. However, adaptively adjusting their contributions usually leads to optimal results. 3) The end-to-end methods (i.e., \ourmethod, OCGIN, and GLocalKD) generally perform better than the two-stage methods. Such an observation illustrates the significance of consistent learning objectives with OOD detection tasks. 4) Among all two-stage methods, GCL methods with Mahalanobis detector demonstrate impressive results in OOD detection. The results show that this competitive solution for OOD detection on vision/language data~\cite{ssd_sehwag2020ssd,nlpood_zhou2021contrastive}, to certain extents, is also useful for graph-structured data. 5) The graph kernel-based methods, unfortunately, do not show a clear advantage over random guessing (AUC$=50\%$). Their performance is possibly limited by: a) they fail to capture feature information; b) they only focus on patterns at the motif level; c) they generate representations and conduct OOD detection separately. 

\begin{table*}[t]
\centering
\caption{Ablation study results of \ourmethodsimp and its variants in terms of AUC (in percent, mean $\pm$ std).} 
\label{tab:ablation}
\vspace{-0.3cm}
\resizebox{1\textwidth}{!}{
\begin{tabular}{ccc | cccccccccc}
\toprule
\multirow{2}{*}{$\mathcal{L}_{node}$} & \multirow{2}{*}{$\mathcal{L}_{graph}$} & \multirow{2}{*}{$\mathcal{L}_{group}$} & BZR & PTC-MR & AIDS & ENZYMES & IMDB-M & Tox21 & FreeSolv & BBBP & ClinTox & Esol \\
\cline{4-13}
  &   &   & COX2 & MUTAG & DHFR & PROTEIN & IMDB-B & SIDER & ToxCast & BACE & LIPO & MUV \\
\hline
\checkmark & -& - & $83.51{\scriptstyle\pm4.14}$ & $72.48{\scriptstyle\pm3.77}$ & $96.84{\scriptstyle\pm0.58}$ & $60.85{\scriptstyle\pm2.95}$ & $79.34{\scriptstyle\pm1.81}$ & $62.58{\scriptstyle\pm0.67}$ & $59.48{\scriptstyle\pm2.20}$ & $69.53{\scriptstyle\pm2.29}$ & $53.29{\scriptstyle\pm4.32}$ & $86.49{\scriptstyle\pm1.20}$  \\
- & \checkmark& - & $87.44{\scriptstyle\pm4.66}$ & $77.84{\scriptstyle\pm3.71}$ & $97.60{\scriptstyle\pm1.05}$ & $56.74{\scriptstyle\pm1.96}$ & $75.22{\scriptstyle\pm1.91}$ & ${65.07{\scriptstyle\pm1.32}}$ & $\mathbf{78.40{\scriptstyle\pm6.44}}$ & $77.66{\scriptstyle\pm2.29}$ & $\underline{70.11{\scriptstyle\pm2.44}}$ & $89.57{\scriptstyle\pm2.80}$  \\
- & -& \checkmark & $79.21{\scriptstyle\pm5.60}$ & $74.83{\scriptstyle\pm8.54}$ & $89.47{\scriptstyle\pm1.85}$ & $50.43{\scriptstyle\pm7.41}$ & $72.91{\scriptstyle\pm2.75}$ & $54.84{\scriptstyle\pm2.56}$ & $58.16{\scriptstyle\pm6.23}$ & $58.09{\scriptstyle\pm5.43}$ & $58.46{\scriptstyle\pm5.35}$ & $83.35{\scriptstyle\pm2.71}$  \\
\checkmark & \checkmark& - & $\mathbf{93.14{\scriptstyle\pm3.63}}$ & $77.53{\scriptstyle\pm4.02}$ & $\underline{98.90{\scriptstyle\pm0.42}}$ & $\underline{61.48{\scriptstyle\pm3.46}}$ & $79.55{\scriptstyle\pm1.35}$ & $\underline{65.44{\scriptstyle\pm1.13}}$ & $71.45{\scriptstyle\pm4.23}$ & $\underline{80.43{\scriptstyle\pm2.57}}$ & $65.89{\scriptstyle\pm4.57}$ & $\underline{90.94{\scriptstyle\pm1.16}}$  \\
\checkmark & -& \checkmark & $85.01{\scriptstyle\pm3.05}$ & $76.10{\scriptstyle\pm3.01}$ & $96.87{\scriptstyle\pm0.52}$ & $59.69{\scriptstyle\pm1.89}$ & $\underline{79.69{\scriptstyle\pm1.67}}$ & $63.01{\scriptstyle\pm0.97}$ & $56.30{\scriptstyle\pm5.33}$ & $69.66{\scriptstyle\pm2.45}$ & $54.14{\scriptstyle\pm4.01}$ & $86.31{\scriptstyle\pm1.99}$  \\
- & \checkmark& \checkmark & $86.59{\scriptstyle\pm5.24}$ & $\underline{77.97{\scriptstyle\pm4.00}}$ & $97.22{\scriptstyle\pm1.35}$ & $55.51{\scriptstyle\pm4.39}$ & $76.17{\scriptstyle\pm1.65}$ & $\mathbf{65.48{\scriptstyle\pm0.78}}$ & $\underline{77.38{\scriptstyle\pm5.19}}$ & $79.77{\scriptstyle\pm4.39}$ & $\mathbf{70.20{\scriptstyle\pm1.01}}$ & $88.33{\scriptstyle\pm1.35}$  \\
\hline
\checkmark & \checkmark& \checkmark & $\underline{93.00{\scriptstyle\pm3.20}}$ & $\mathbf{78.43{\scriptstyle\pm2.67}}$ & $\mathbf{98.91{\scriptstyle\pm0.41}}$ & $\mathbf{61.89{\scriptstyle\pm2.51}}$ & $\mathbf{79.71{\scriptstyle\pm1.19}}$ & $65.30{\scriptstyle\pm1.27}$ & $70.48{\scriptstyle\pm2.75}$ & $\mathbf{81.56{\scriptstyle\pm1.97}}$ & $66.13{\scriptstyle\pm2.98}$ & $\mathbf{91.39{\scriptstyle\pm0.46}}$  \\
\bottomrule
\end{tabular}}
\vspace{-0.1cm}
\end{table*}

\vspace{-1.5mm}
\subsection{Performance on Anomaly Detection (RQ2)}
To investigate if \ourmethod can generalize to anomaly detection setting~\cite{ocgin_zhao2021using,glocalkd_ma2022deep}, we conduct anomaly detection experiments on 15 datasets following the benchmark in \cite{glocalkd_ma2022deep}. The results are illustrated in Table~\ref{tab:main_result_ad}. From the results, we find that our proposed methods also perform well in anomaly detection settings. The main reason is that \ourmethod captures common patterns in three different scale levels, leading to its strong power in modeling normal data. In contrast, the baseline methods only consider one or two scale levels, resulting in sub-optimal performance. Similar to the observations in Section \ref{subsec:od_results}, we can also find that the end-to-end methods generally outperform the two-stage methods, and the kernel-based methods tend to perform worse than other baselines. These observations show the effectiveness of some key designs in our methods, i.e., end-to-end training and feature/structure views construction.

\vspace{-1.5mm}
\subsection{Ablation Study (RQ3)}

Our methods consider hierarchical graph contrastive learning with contrasts in three levels, i.e., node level, graph level, and group level. To verify the effectiveness of each component, we conduct experiments on all combinations of them. To eliminate the influence of adaptive training and scoring mechanism, we perform the ablation study on \ourmethodsimp that equally combines three components via unweighted summation. The experimental results on our OOD detection benchmark are reported in Table~\ref{tab:ablation}, which brings the following observations. First, \ourmethodsimp that uses all components (the last row) achieves the best results on 6 out of 10 datasets, and has promising performance on the rest datasets. This observation indicates the effectiveness of jointly executing contrastive learning of multiple graph levels for OOD detection. Second, contrastive learning at each level brings considerable contribution, while graph-level contrast generally contributes more. This observation verifies the effectiveness of each component. Third, compared to considering an individual component, combining the contrasts at two levels usually improves the performance. The possible reason is that contrasts at different levels would expose the OOD patterns from different perspectives, leading to more comprehensive detection performance. Fourth, on some datasets, directly adding the loss/score terms may lead to sub-optimal results, which illustrates the significance of introducing an adaptive mechanism. Taking dataset pair FreeSolv/ToxCast as an example, the performance of \ourmethodsimp ($70.48\%$) is lower than which of only using graph-level contrast ($78.40\%$); by considering the adaptive mechanism in \ourmethod, the AUC can increase to $80.13\%$.

\vspace{-1.5mm}
\subsection{Parameter Study (RQ4)} \label{subsec:param}

\begin{figure}[t!]
 \centering
 \vspace{-0.33cm}
 \subfigure[Sensitivity of $K$]{
   \includegraphics[width=0.18\textwidth]{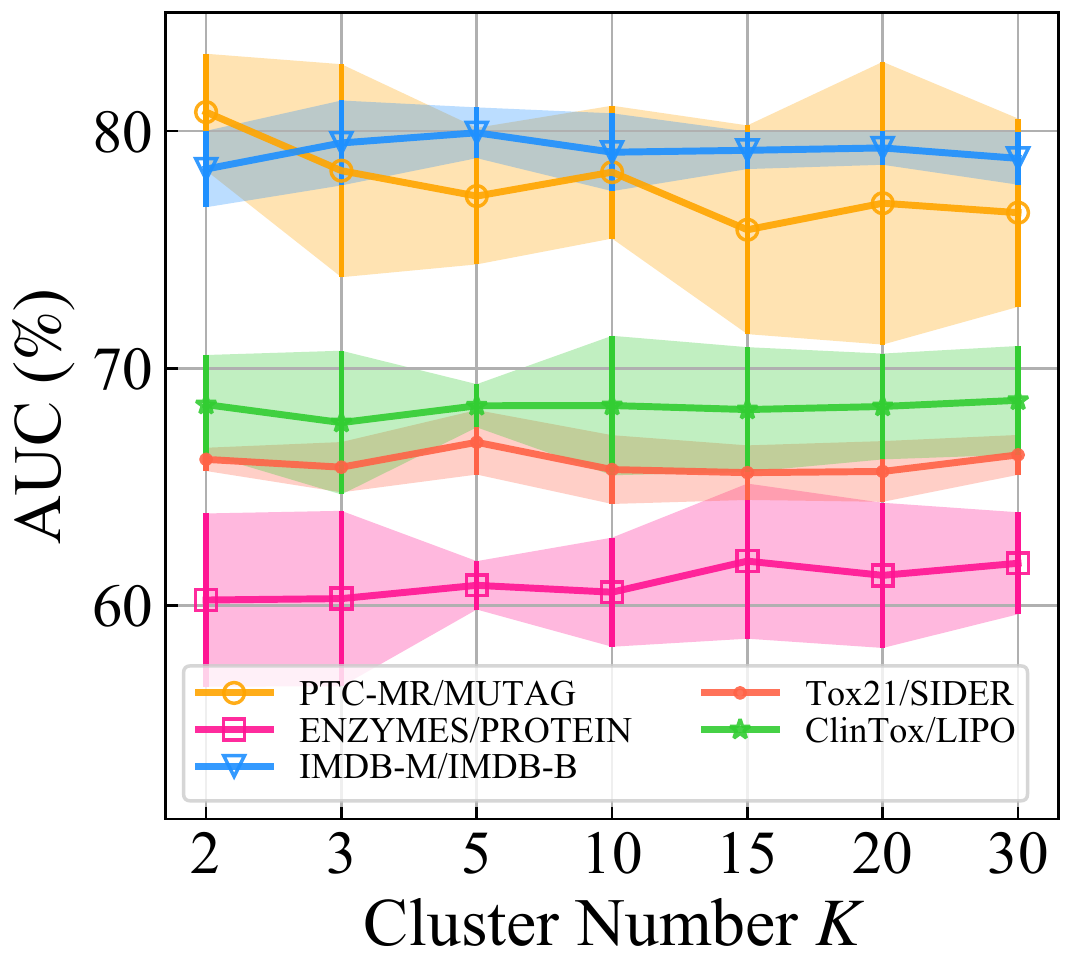}
   \label{subfig:param_k}
 } 
 \hspace{0.3cm}
 \subfigure[Sensitivity of $\alpha$]{
   \includegraphics[width=0.18\textwidth]{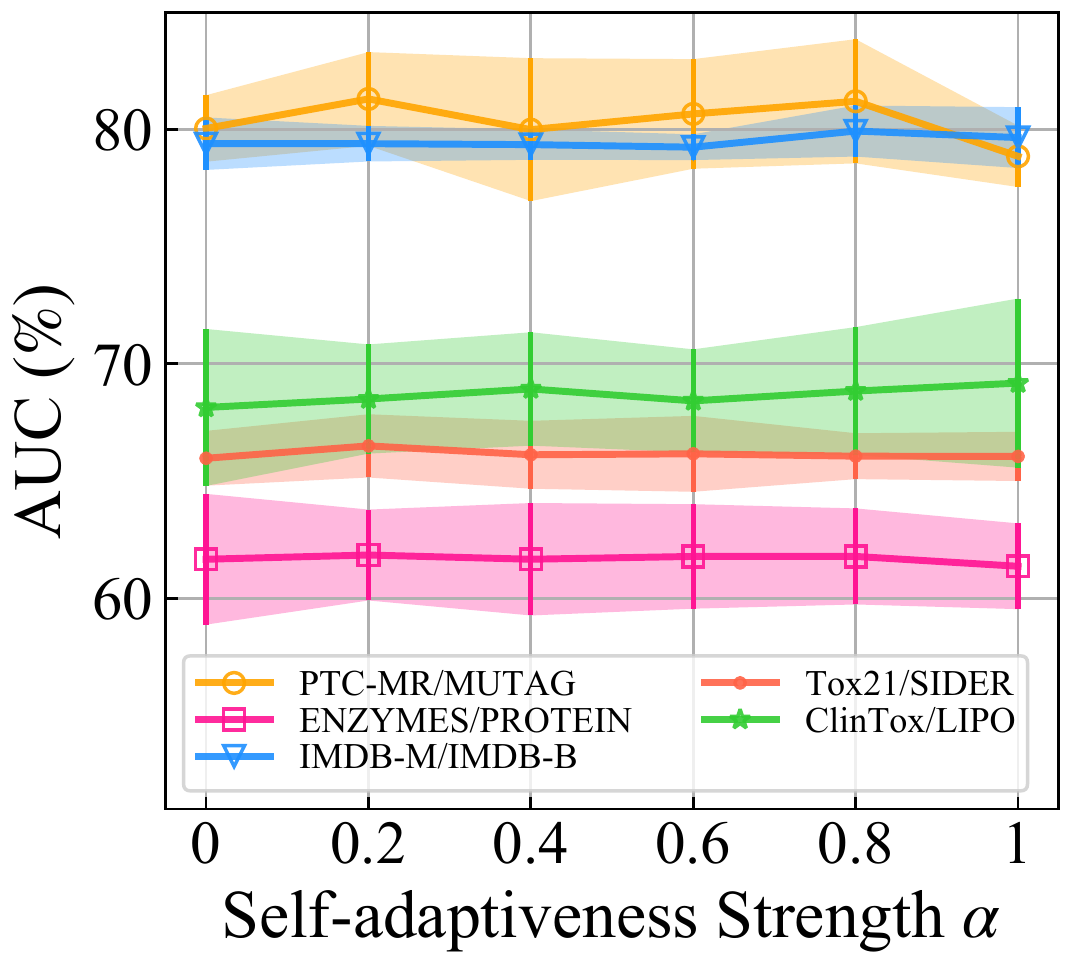}
   \label{subfig:param_a}
 }
 \vspace{-0.4cm}
 \caption{Parameter sensitivity of $K$ and $\alpha$.}
 \vspace{-0.3cm}
 \label{fig:param_ak}
\end{figure}

\noindent \textbf{Cluster number $K$.} 
We study the sensitivity of \ourmethod w.r.t. the cluster number $K$ by varying $K$ as $\{2, 3, 5, 10, 15, 20, 30\}$. As shown Fig.~\ref{subfig:param_k}, the best selection of $K$ for different dataset pairs is quiet different. For instance, PTC-MR/MUTAG needs fewer clusters ($K=2$), while a larger $K$ is preferred by ENZYMES/PROTEIN. We conjecture that the best selection of $K$ is highly related to the properties of ID datasets, such as the number of categories. Fortunately, \ourmethod is not very sensitive to this hyper-parameter, and a moderate value (i.e., $K=5,10,15$) usually result in respectable performance. \looseness-1

\noindent \textbf{Self-adaptiveness strength $\alpha$.} 
To analyze the sensitivity of $\alpha$ for \ourmethod, we alter the value of $\alpha$ from $0$ to $1$. The AUC w.r.t different selection of $\alpha$ is plotted in Fig.~\ref{subfig:param_a}. From the figure, we can find that the AUC would drop slightly when $\alpha=0$, illustrating the significance of self-adaptive mechanism for loss function. In general, the performance is relatively stable across different values of $\alpha$, and the best results often occur when $\alpha$ is between $0.2$ and $0.8$. \looseness-1

\vspace{-1.5mm}
\subsection{Visualization (RQ5)}

\begin{figure}[t!]
 \centering
 \vspace{-0.33cm}
 \subfigure[Node-space feat. emb.]{
   \includegraphics[width=0.15\textwidth]{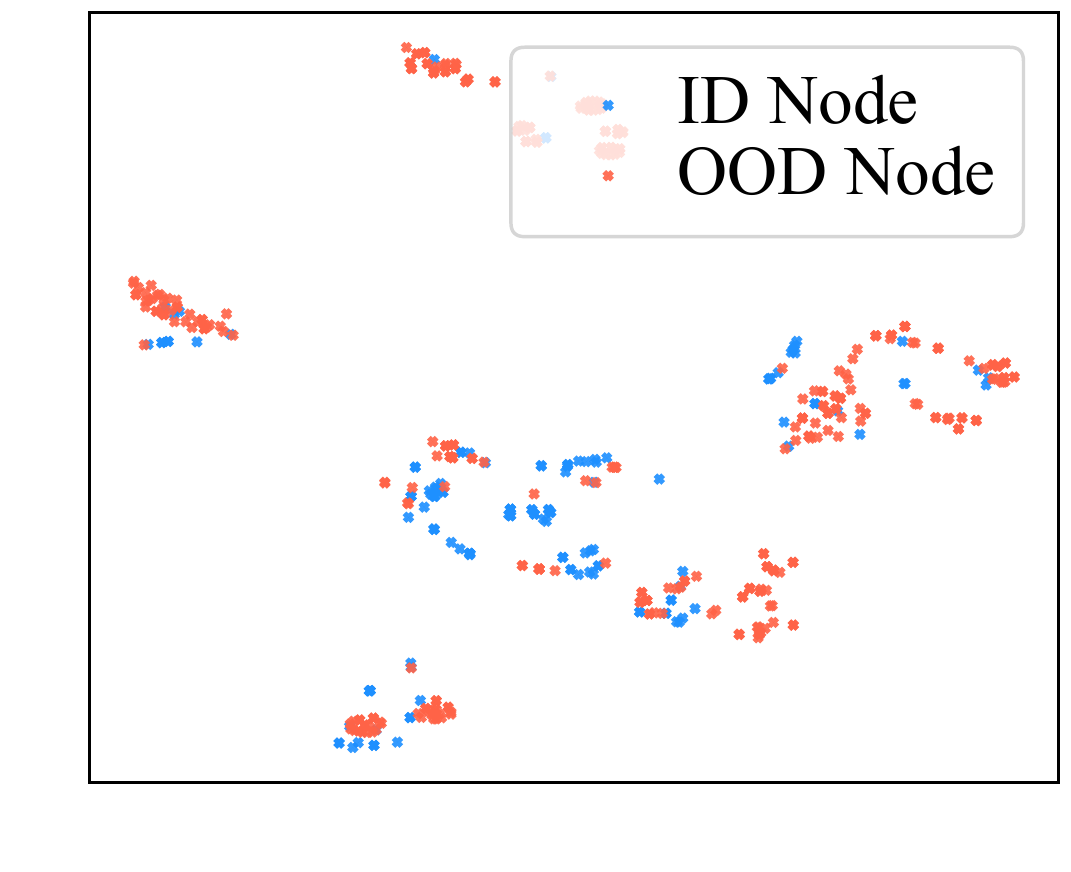}
   \label{subfig:node_f}
 } 
 \hspace{-0.18cm}
 \subfigure[Node-space str. emb.]{
   \includegraphics[width=0.15\textwidth]{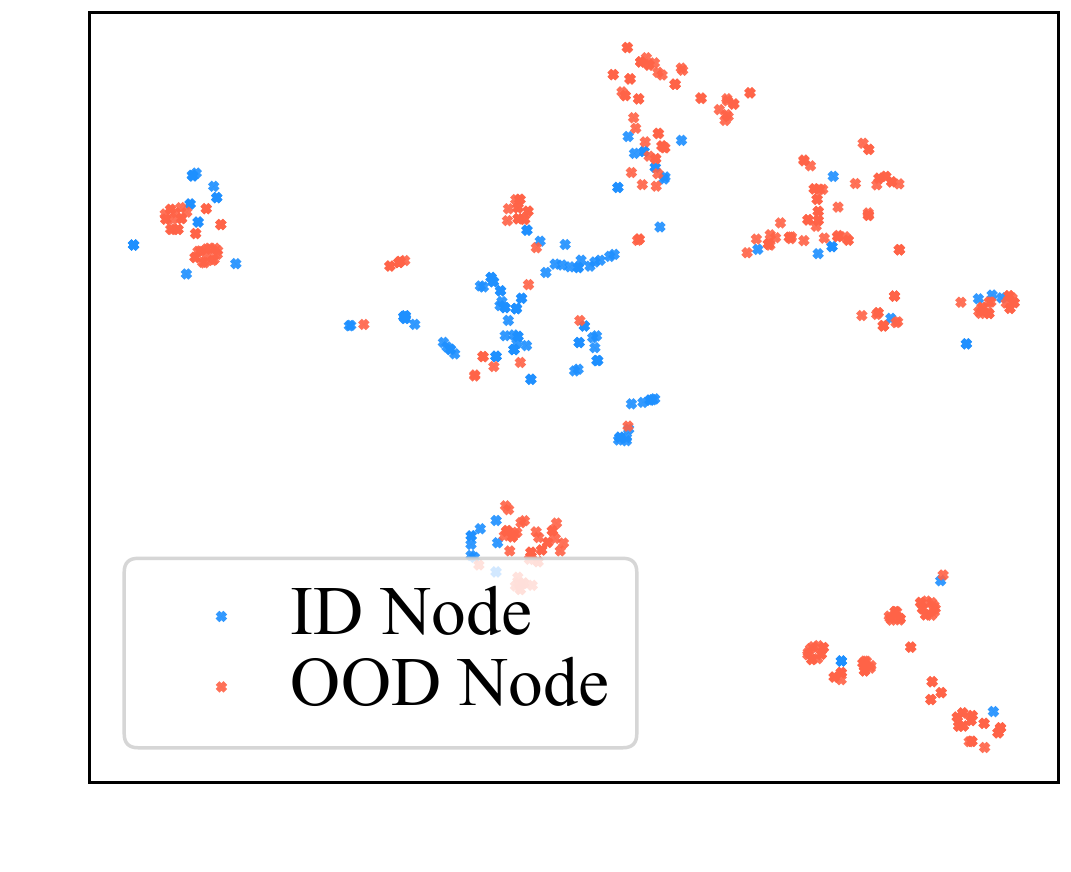}
   \label{subfig:node_s}
 } 
 \hspace{-0.18cm}
 \subfigure[Graph-space feat. emb.]{
   \includegraphics[width=0.15\textwidth]{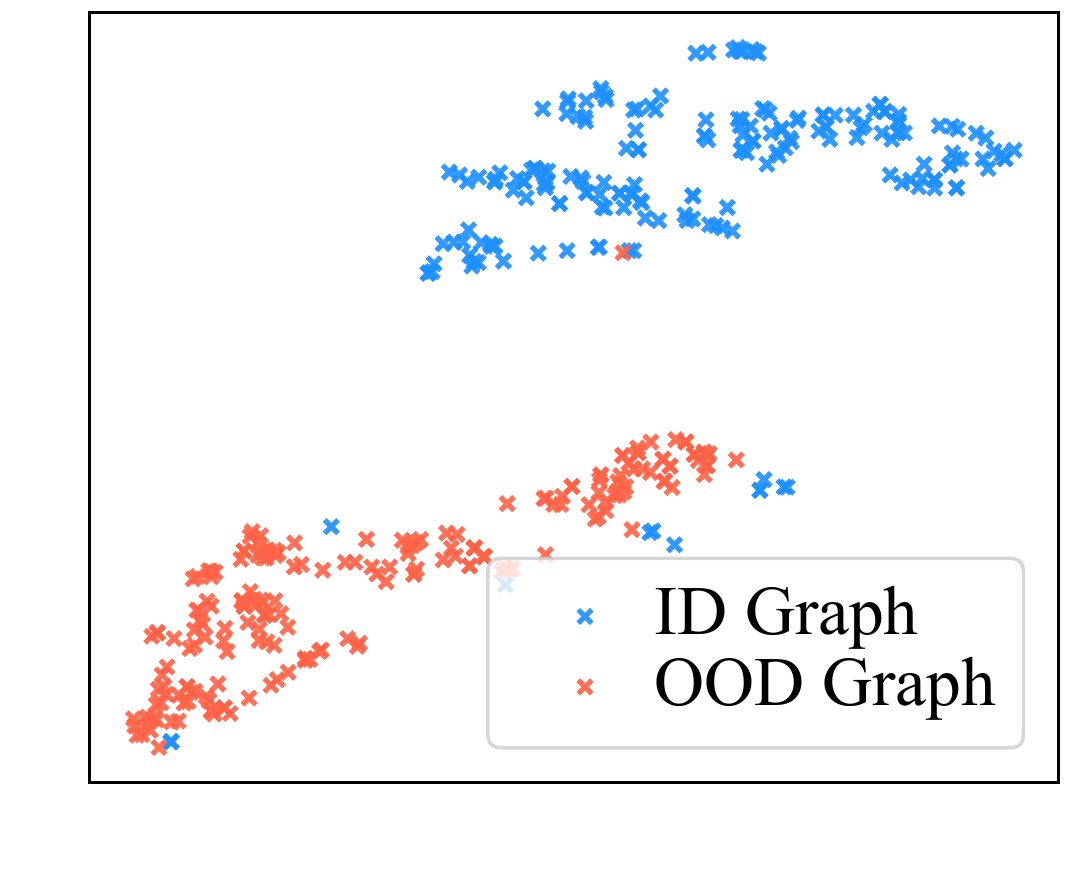}
   \label{subfig:graph_f}
 } 
 \subfigure[Graph-space str. emb.]{
   \includegraphics[width=0.15\textwidth]{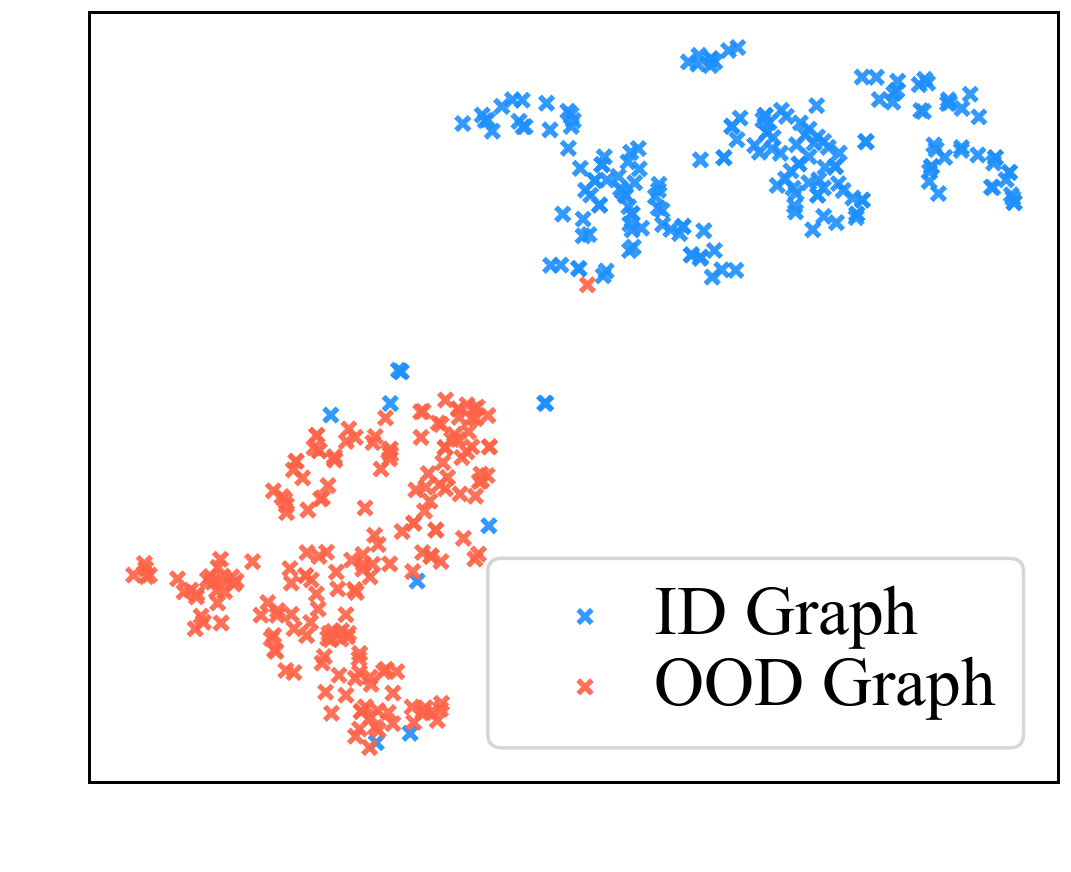}
   \label{subfig:graph_s}
 } 
 \hspace{-0.18cm}
 \subfigure[Group-space emb.]{
   \includegraphics[width=0.15\textwidth]{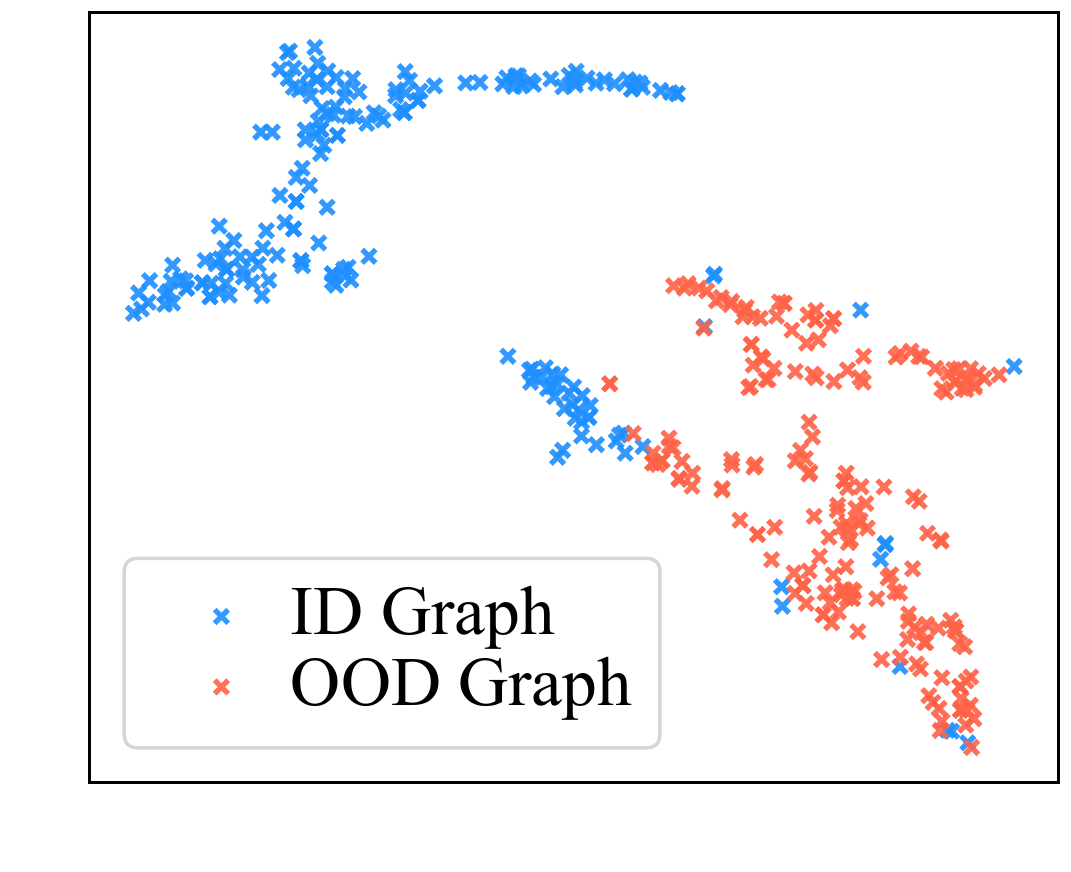}
   \label{subfig:group}
 } 
 \hspace{-0.18cm}
 \subfigure[OOD score]{
   \includegraphics[width=0.15\textwidth]{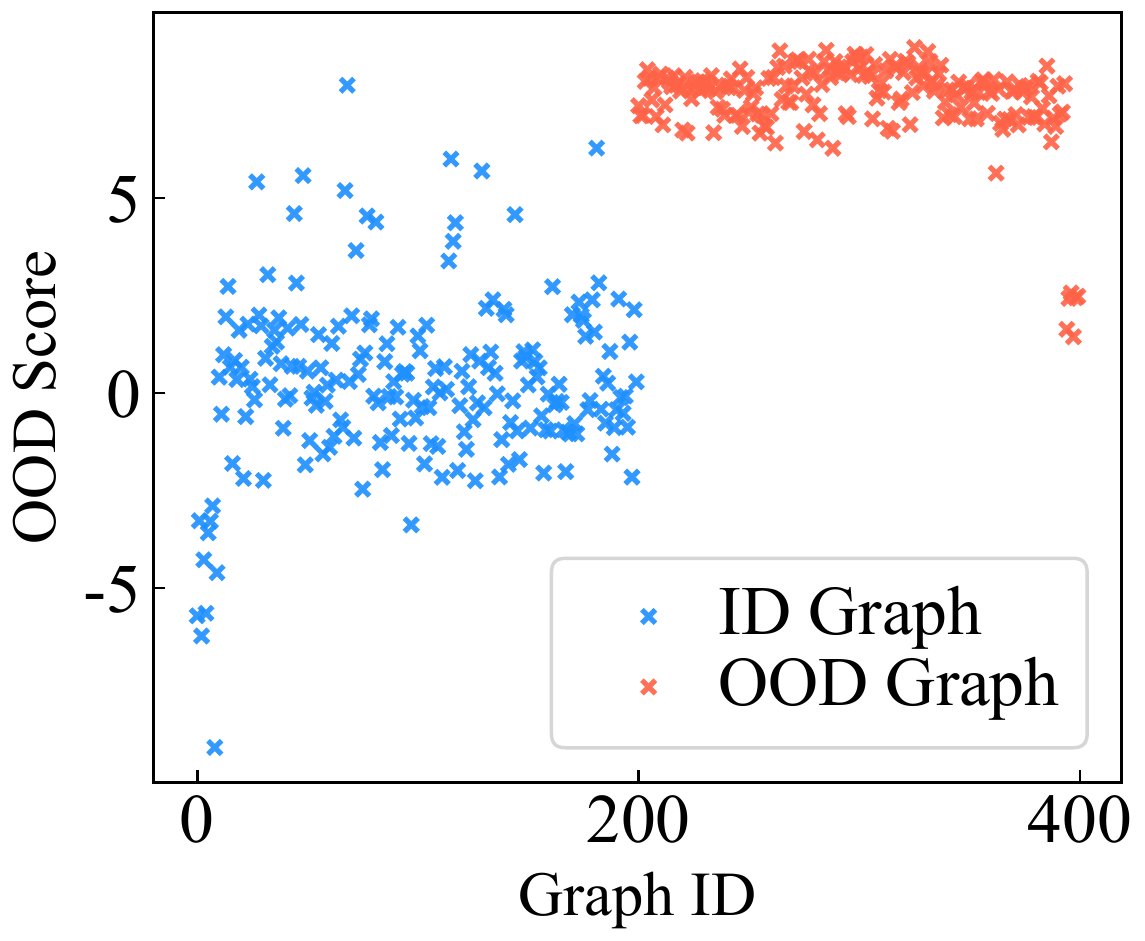}
   \label{subfig:score}
 } 
 \vspace{-0.3cm}
 \caption{Visualization on AIDS/DHFR dataset pair. (a)-(e): t-SNE visualization of testing sample embeddings (emb.) of feature (feat.) or structure (str.) view at different embedding spaces. (f): OOD scores of \ourmethod on testing samples.}
 \vspace{-0.11cm}
 \label{fig:visualization}
\end{figure}

To answer RQ5, we use t-SNE~\cite{tsne_van2008visualizing} to visualize the embeddings learned by \ourmethod at different spaces. Fig.~\ref{fig:visualization}(a)-(e) show that the ID samples and OOD samples are well separated in each embedding space. We also visualize the distribution of OOD scores learned by \ourmethod in Fig.~\ref{subfig:score}. We can observe that the OOD samples tend to have OOD scores that are greater than $7$, while the ID samples are given smaller scores ($s<4$). Such a clear scoring boundary leads to the superior OOD detection performance of \ourmethod. 

\section{Conclusion}
In this paper, we make the first attempt toward detecting out-of-distribution (OOD) samples from graph-structured data. To tackle this problem, we propose a novel OOD detection method termed \ourmethod, which learns the attributive and structural patterns from training in-distribution (ID) by a carefully-crafted hierarchical graph contrastive learning framework. In \ourmethod, the contrasts at node, graph, and group levels are jointly conducted by maximizing the mutual agreement between feature and structure graph view, and a self-adaptive mechanism is designed to balance the trade-off among the learning objectives and learned OOD scores at three levels. Extensive experiments demonstrate the superiority of \ourmethod over the baseline methods in a series of real-world benchmarks. 

\begin{acks}
This work was supported by ARC Future Fellowship (No. FT210100097), Amazon Research Award, ONR (No. N00014-21-1-4002), ARO (No. W911NF2110030), ARL (No. W911NF2020124), and NSF (No. 2229461).
\end{acks}

\bibliographystyle{ACM-Reference-Format}
\bibliography{ref}

\end{document}